\documentclass{article}

\usepackage[preprint]{neurips_2026}

\usepackage[utf8]{inputenc} 
\usepackage[T1]{fontenc}    
\usepackage{hyperref}       
\usepackage{url}            
\usepackage{booktabs}       
\usepackage{amsfonts}       
\usepackage{nicefrac}       
\usepackage{microtype}      
\usepackage{xcolor} 
\usepackage{amsmath}
\usepackage{graphicx}

\usepackage[utf8]{inputenc} 
\usepackage[T1]{fontenc}    
\usepackage{url}            
\usepackage{booktabs}       
\usepackage{amsfonts}       
\usepackage{nicefrac}       
\usepackage{microtype}      
\usepackage[dvipsnames,table]{xcolor}
\usepackage{wrapfig}
\usepackage[ruled,vlined]{algorithm2e}
\usepackage{cancel}
\usepackage{caption}
\usepackage{graphicx}
\usepackage{capt-of}
\usepackage{graphicx}
\usepackage{tabularx}
\usepackage{subcaption}

\usepackage{amsmath,amssymb}

\usepackage{booktabs}
\usepackage[table]{xcolor}
\usepackage{array}

\usepackage{adjustbox}

\definecolor{tabgreen}{HTML}{DFF2E1}
\definecolor{tabamber}{HTML}{FFF1CC}
\definecolor{tabred}{HTML}{F8DDDD}
\definecolor{tabsection}{HTML}{F5F5F5}

\usepackage{booktabs}
\usepackage{multirow}
\usepackage[table]{xcolor}

\definecolor{lightgray}{HTML}{eeeeee}

\newcommand{\fullgray}[1]{%
  \multicolumn{1}{>{\columncolor{lightgray}}c}{#1}%
}

\usepackage{xcolor}
\usepackage{listings}
\usepackage{caption}
\usepackage{wrapfig}

\definecolor{algocomment}{RGB}{70,130,135}
\definecolor{algopink}{RGB}{220,45,120}
\definecolor{algoblue}{RGB}{40,60,255}

\lstdefinelanguage{trainalgo}{
  sensitive=true,
  morecomment=[l]{\#},
  commentstyle=\color{algocomment},
  alsoletter={_},
  morekeywords={sample_t,randn_like,net,l2_loss},
  keywordstyle=\color{algopink},
  literate=
    {*}{{{\color{algoblue}*}}}1
    {+}{{{\color{algoblue}+}}}1
    {-}{{{\color{algoblue}-}}}1
    {/}{{{\color{algoblue}/}}}1,
}

\usepackage{xcolor}
\usepackage{listings}

\definecolor{algocomment}{RGB}{80,140,140}   
\definecolor{algopink}{RGB}{220,50,120}      
\definecolor{algogreen}{RGB}{0,95,45}        

\lstdefinelanguage{trainalgo}{
  sensitive=true,
  alsoletter={_},
  morecomment=[l]{\#},
  commentstyle=\color{algocomment}\bfseries,
  morekeywords=[1]{sample_t,randn_like,rand,drop_labels,net,l2_loss,split,RMSNorm,concat,chunk,silu,Linear,LoRA},
  keywordstyle=[1]\color{algopink}\bfseries,
  morekeywords=[2]{use_regs, regs, w_rg, p},
  keywordstyle=[2]\color{MidnightBlue}\bfseries,
}

\newcounter{snippet}
\renewcommand{\thesnippet}{\arabic{snippet}}

\newcommand{\snippettitle}[2]{%
  \refstepcounter{snippet}%
  \label{#1}%
  {\bfseries \footnotesize Snippet~\thesnippet. #2}%
}
\usepackage{makecell}

\newcommand{\cfg}[1]{\makecell[l]{
${t}_{\min}^{\mathrm{cfg}}=0.1$\\[1.4mm]
$t_{\max}^{\mathrm{cfg}}=1.0$\\[1.4mm]
$w_{\mathrm{cfg}}=#1$
}}

\newcommand{\rg}[1]{\makecell[l]{
$t_{\min}^{\mathrm{rg}}=0.03$\\[1.4mm]
$t_{\max}^{\mathrm{rg}}=0.9$\\[1.4mm]
$w_{\mathrm{rg}}=#1$
}}

\newcommand{\rgpart}[1]{\makecell[l]{
$t_{\min}^{\mathrm{rg}}=0.1$\\[1.4mm]
$t_{\max}^{\mathrm{rg}}=1.0$\\[1.4mm]
$w_{\mathrm{rg}}=#1$
}}

\newcommand{\cfgpart}[1]{\makecell[l]{
$t_{\min}^{\mathrm{cfg}}=0.1$\\[1.4mm]
$t_{\max}^{\mathrm{cfg}}=1.0$\\[1.4mm]
$w_{\mathrm{cfg}}=#1$
}}

\title{Registers Matter for Pixel-Space \\ Diffusion Transformers}

\author{%
Nikita Starodubcev \quad
Ilia Sudakov\thanks{Equal contribution.} \quad
Ilya Drobyshevskiy\footnotemark[1] \\
\textbf{Artem Babenko} \quad
\textbf{Dmitry Baranchuk} \\[0.5em]
Yandex Research \\ \\
\makebox[\textwidth][c]{\href{https://quickjkee.github.io/registers-project-page/}{\texttt{Project page}}}
}

\begin{document}

\maketitle

\begin{abstract}
Vision Transformers (ViTs) are known to exhibit high-norm patch-token outliers that degrade feature map quality, a problem effectively mitigated by \textit{register tokens}. 
As diffusion models increasingly adopt transformer architectures and move toward pixel-space training, they become closer in form to ViTs, raising the question of whether register tokens are also useful for Diffusion Transformers (DiTs).
In this work, we show that DiTs differ from ViTs in a key respect: \textit{they do not exhibit patch-token outliers but still benefit from registers.} Interestingly, registers are more effective in pixel-space DiTs than in latent-space DiTs.
By analyzing intermediate representations, we find that register tokens produce cleaner feature maps at high noise levels, which may contribute to their effectiveness in pixel-space generation. 
We further observe that recent pixel-space DiT architectures implicitly incorporate register-like mechanisms, which may partially account for their strong empirical performance. Motivated by these observations, we propose \textit{Register Guidance}, a technique that amplifies the contribution of register tokens responsible for improving visual structure and coherence.
\end{abstract}

\section{Introduction}

Vision Transformers (ViTs)~\cite{dosovitskiy2020image, liu2021swin, touvron2021training} have become a dominant architecture for visual representation learning by modeling images as sequences of patch tokens processed via self-attention~\cite{vaswani2017attention}.
Recent advances in self-supervised learning (SSL)~\cite{caron2021emerging,oquab2023dinov2,simeoni2025dinov3} demonstrate that ViTs trained on unlabeled data can learn semantically meaningful representations, enabling object- and part-level understanding useful for downstream tasks such as unsupervised segmentation and detection~\cite{simeoni2021localizing,hamilton2022unsupervised,amir2021deep,oquab2023dinov2,wang2023tokencut}.

Recent research has focused on understanding the emergence of high-norm tokens in ViTs, which are often associated with artifacts in attention maps~\cite{darcet2023vision,jiang2025vision,lappe2025register,shi2026vision,chen2025vision,wang2024sinder}.
As these artifacts lead to less interpretable attention maps and weaker performance on dense prediction tasks, \citep{darcet2023vision} proposes using additional \textit{register tokens} to prevent patch tokens from being repurposed for global representations.

In parallel, diffusion models (DMs)~\cite{ho2020denoising,song2019generative} have widely adopted transformer-based architectures~\cite{peebles2023scalable,ma2024sit}, replacing convolutional backbones~\cite{ronneberger2015u,dhariwal2021diffusion}.
Recent work has also revisited training directly in pixel space~\cite{li2025back,yu2025pixeldit,lu2026one}, as an alternative to latent diffusion models that rely on pretrained autoencoders~\cite{rombach2022high,podell2023sdxl}. This progress brings Diffusion Transformers (DiTs) closer to ViTs
and motivates two questions: \textit{(1) do DiTs inherit high-norm patch-token outliers similar to those observed in ViTs?} and \textit{(2) can register tokens also be effective in these models?}

These questions are also related to broader studies of attention sinks and special-token behavior in transformers~\cite{su2026attention}, including generative models~\cite{xiao2023efficient,gu2024attention,rulli2025attention,liu2025rolling,jamal2026diffusion}. 
We refer a detailed discussion of this direction to App.~\ref{app:related} and focus here on image diffusion transformers, where the presence and role of register-like tokens remain underexplored.

\textbf{Contributions.} 
We find that, unlike ViTs, diffusion transformers in both latent and pixel spaces do not exhibit noticeable high-norm outliers among patch tokens. 
Interestingly, despite the absence of such outliers, adding register tokens to DiTs leads to the emergence of high-norm tokens within the registers themselves.

Importantly, we observe that the impact of register tokens differs across training spaces: pixel-space models benefit the most, whereas latent-space models show only moderate gains or even slightly degraded performance.

Accordingly, we focus our study on pixel-space DiTs and find that registers benefit them through mechanisms distinct from those in ViTs.
Our analysis shows that registers consistently reduce patch-token feature norms and produce smoother intermediate feature maps, especially at high noise levels. 
Moreover, register tokens specialize into distinct roles: some act as norm sinks, while others encode global semantic information.

These findings also provide a rationale for recent pixel-space DiT designs~\cite{li2025back, lu2026one}, which introduce additional in-context class-conditioning and achieve substantial performance gains. 
Our analysis suggests that these gains may largely arise from register-like behavior rather than from additional class information. 
In particular, in-context tokens behave similarly to register tokens, with some encoding global semantic information and others acting as norm sinks. We observe a similar phenomenon in the text-to-image setting, where some tokens from the text sequence also become sinks.

From a practical standpoint, we exploit the observation that registers improve object structure and coherence while preserving the main content. This motivates \emph{Register Guidance}, which treats the prediction of a model without registers as a weak model~\cite{karras2024guiding} and uses it as the negative direction. Combined with CFG~\cite{ho2022classifier}, RG consistently improves the performance of recent pixel-space DiTs.

\section{Register Tokens for Image Diffusion Transformers}

\begin{figure*}[t!]
    \centering
\includegraphics[width=\textwidth]{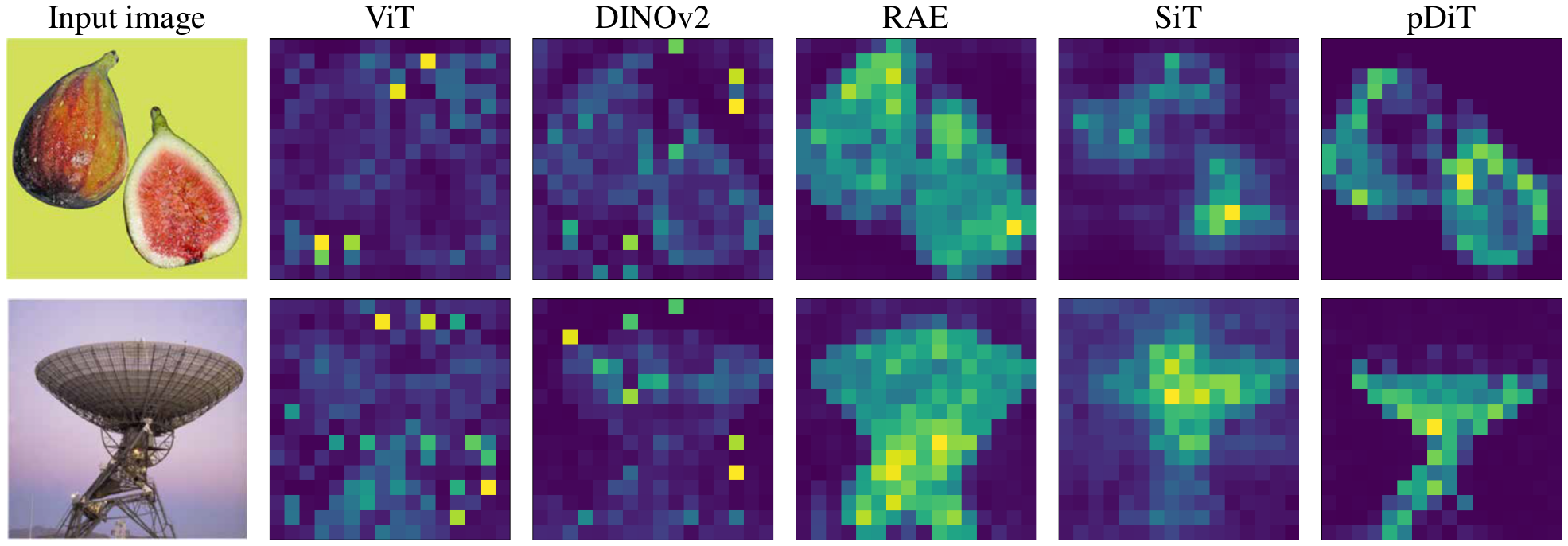}
    \caption{
        \textbf{Diffusion transformers do not exhibit attention-map outliers.} 
        Unlike ViTs, where attention-map anomalies typically appear in low-information regions (e.g., background), DiT attention remains largely focused on the main objects.
    }
    \label{fig:analysis_attn_maps_dino_vs_jit}
\end{figure*}

In this section, we analyze the role of register tokens in image DiTs and highlight key differences from their use in ViTs. 
As a representative ViT-based model, we consider DINOv2~\cite{oquab2023dinov2}.
 
For diffusion models, we primarily focus on pixel-space DiTs based on the standard architecture~\cite{peebles2023scalable} with widely used transformer improvements~\cite{yao2025reconstruction, li2025back}; we refer to these models as pDiTs. 
We train pDiTs using flow matching~\cite{lipman2023flow, albergo2025stochastic} on ImageNet~\cite{deng2009imagenet} at resolution $256{\times}256$ with patch size $16$. We also consider the more advanced pixel-space architecture PixelDiT~\cite{yu2025pixeldit}, following the original setup.

We additionally analyze latent-space architectures, SiT~\cite{ma2024sit} and RAE~\cite{zheng2026diffusion}, using their original backbone designs and training pipelines. 
For RAE, we use a DINOv2-based model~\cite{oquab2023dinov2}. 
Generation quality is evaluated using FID~\cite{heusel2017gans}.

We study models of varying sizes, with and without register tokens. 
Registers are implemented as additional learnable tokens appended to the patch-token sequence following~\cite{darcet2023vision} and are not used in the training objective. 
Further details are provided in App.~\ref{app:implementation_details}.

\subsection{Registers Benefit Diffusion Transformers Despite the Absence of Outliers}

\begin{figure*}[t!]
    \centering
\includegraphics[width=\textwidth]{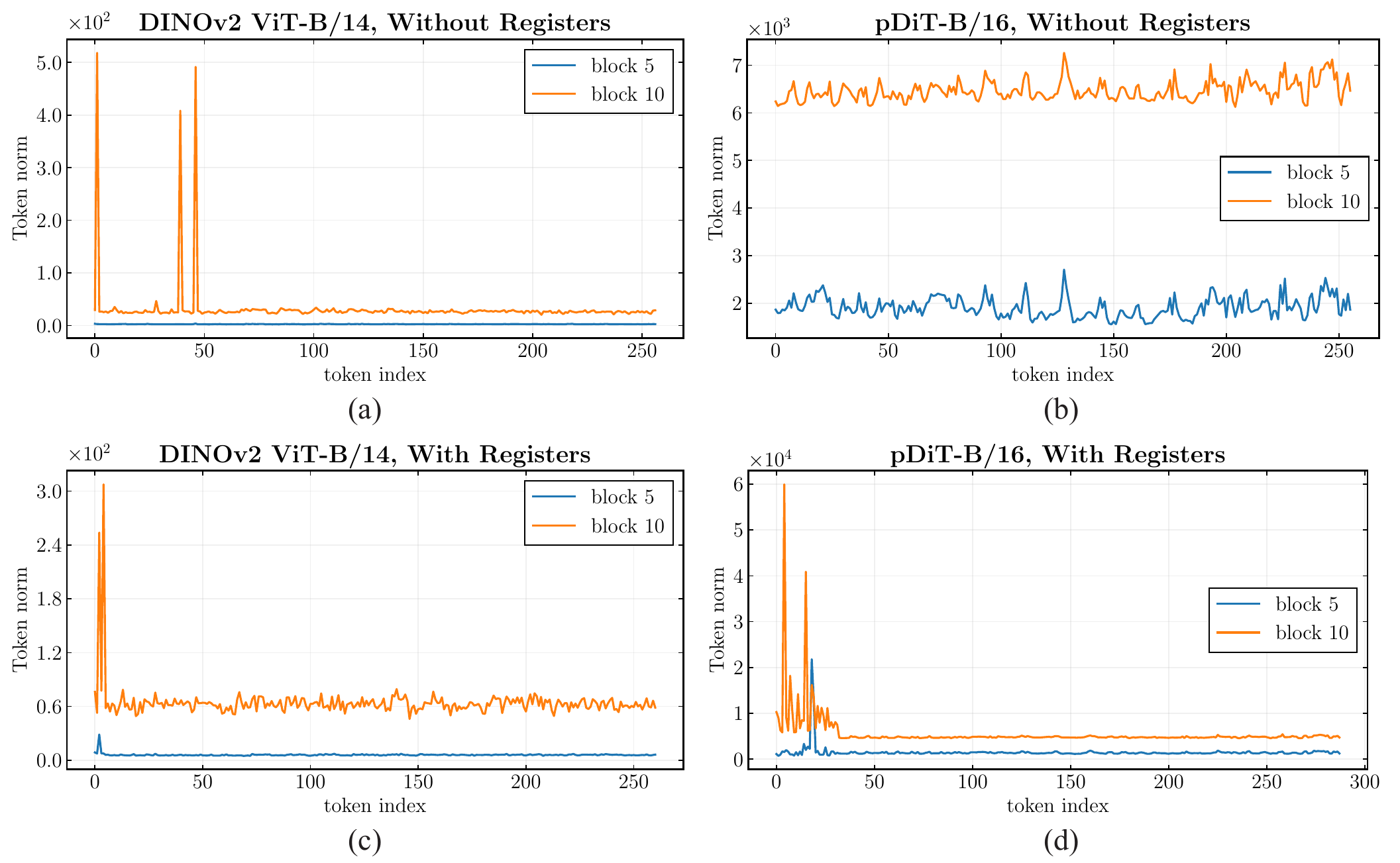}
    \caption{
        \textbf{Token-norm behavior in DINOv2 and pDiTs. Top: without registers; bottom: with registers.}
        (a) In DINOv2, anomalies are localized to a few image tokens with significantly higher norms than the rest.
        (b) In contrast, pDiTs show no patch-token norm outliers, suggesting that registers may be unnecessary in this case.
        (c) As expected, adding register tokens to DINOv2 shifts high-norm outliers into these tokens.
        (d) Interestingly, pDiTs also develop high-norm register tokens, even though such outliers are absent without registers.
    }
\label{fig:analysis_dino_vs_jit_with_and_wo_regs}
\end{figure*}


\begin{table*}[t!]
\centering
\footnotesize
\renewcommand{\arraystretch}{1.12}

\newcommand{\noreg}{\shortstack{w/o regs.}}
\newcommand{\wreg}{\shortstack{w/ regs.}}

\begin{minipage}[t]{0.70\textwidth}
\vspace{0pt}
\centering
\setlength{\tabcolsep}{0pt}

\begin{tabular*}{\linewidth}{@{\extracolsep{\fill}}lccccccccc@{}}
\toprule[1.05pt]
& \multicolumn{3}{c}{pDiT-B/16}
& \multicolumn{3}{c}{pDiT-L/16}
& \multicolumn{3}{c}{pDiT-H/16} \\
\cmidrule(lr){2-4}
\cmidrule(lr){5-7}
\cmidrule(lr){8-10}
Epoch
& \noreg & \wreg & \fullgray{IC}
& \noreg & \wreg & \fullgray{IC}
& \noreg & \wreg & \fullgray{IC} \\
\midrule
$200$
& $7.39$ & $\mathbf{5.30}$ & \fullgray{$4.71$}
& $4.13$ & $\mathbf{3.17}$ & \fullgray{$2.95$}
& $3.52$ & $\mathbf{2.69}$ & \fullgray{$2.38$} \\

$600$
& $4.80$ & $\mathbf{3.80}$ & \fullgray{$3.71$}
& $2.80$ & $\mathbf{2.47}$ & \fullgray{$2.47$}
& $2.35$ & $\mathbf{2.02}$ & \fullgray{$1.90$} \\
\bottomrule[1.05pt]
\end{tabular*}
\end{minipage}%
\hfill
\begin{minipage}[t]{0.27\textwidth}
\vspace{0pt}
\centering
\setlength{\tabcolsep}{0pt}

\begin{tabular*}{\linewidth}{@{\extracolsep{\fill}}lccc@{}}
\toprule[1.05pt]
& \multicolumn{3}{c}{PixelDiT-XL/16} \\
\cmidrule(lr){2-4}
Epoch
& \noreg & \wreg & \fullgray{IC} \\
\midrule
$80$
& $2.74$ & $\mathbf{2.24}$ & \fullgray{$2.28$} \\

$120$
& $2.25$ & $\mathbf{2.12}$ & \fullgray{${2.12}$} \\
\bottomrule[1.05pt]
\end{tabular*}
\end{minipage}

\vspace{2mm}

\caption{
\textbf{Generation quality (FID $\downarrow$) of pDiTs and PixelDiT~\cite{yu2025pixeldit} with and without register tokens.}
We compare generation quality across model sizes and training epochs.
Here, ``w/o regs.'' and ``w/ regs.'' denote training without and with register tokens, respectively.
\textcolor[HTML]{808080}{The shaded IC column reports performance with in-context class conditioning (Section~\ref{sec:incontext}).}
}
\label{tab:analysis_jit_registers}
\end{table*}






The motivation for introducing register tokens in ViTs is to mitigate outliers in feature maps. 
These outliers manifest as high-norm tokens, often localized in low-information regions, e.g., background.

We first investigate whether such outliers arise in DiTs without registers for different spaces by comparing their attention maps to those of ViTs.
As shown in Figure~\ref{fig:analysis_attn_maps_dino_vs_jit}, DiTs do not exhibit the artifacts observed in ViTs. 
In particular, their attention maps remain free of anomalies in low-information regions, which in ViTs are typically associated with unusually high-norm tokens.

This observation is further supported by Figure~\ref{fig:analysis_dino_vs_jit_with_and_wo_regs}(top), which reports token-wise feature norms across layers for pDiT. 
In contrast to DINOv2, where a few tokens attain significantly larger norms, pDiTs exhibit a nearly uniform distribution of patch-token norms. 
As shown in Figures~\ref{fig:app_analysis_jit_wo_w_regs}, \ref{fig:app_analysis_sit_wo_w_regs}, \ref{fig:app_analysis_rae_wo_w_regs} this behavior consistently holds across larger model variants and latent-space architectures.

Based on these observations, DiTs would not be expected to benefit from register tokens, as the feature map artifacts that originally motivated their use are absent. 
However, contrary to this expectation, we observe the opposite.

First, in Figure~\ref{fig:analysis_dino_vs_jit_with_and_wo_regs}(bottom), we compare token-wise feature norms for DINOv2 and pDiTs with register tokens. 
As expected, in DINOv2, registers primarily absorb pre-existing outliers from patch tokens. 
In contrast, pDiTs develop high-norm tokens within the registers, despite the absence of such outliers in models without registers. 
Figures~\ref{fig:app_analysis_jit_w_regs_more_t} and~\ref{fig:app_analysis_jit_wo_w_regs} show that this effect consistently holds across different timesteps and model sizes. It also extends to latent-space models (Figures~\ref{fig:app_analysis_sit_wo_w_regs} and~\ref{fig:app_analysis_rae_wo_w_regs}) and few-step models (Figures~\ref{fig:app_imf_analysis} and~\ref{fig:app_pmf_analysis}). In App.~\ref{app:analysis_imagenet}, we show that outliers mainly form after the MLP layer.

Second, as shown in Table~\ref{tab:analysis_jit_registers}, introducing register tokens in pDiTs consistently improves generation quality across model sizes. We observe the same effect for PixelDiT and consider a higher resolution in Table~\ref{tab:app_analysis_jit_registers_512}.
However, Table~\ref{tab:analysis_register_spaces} shows that registers provide significantly smaller gains in latent-space models, a phenomenon we discuss in Section~\ref{sec:pixel_space}.

Based on the observations, pDiTs benefit from outliers but lack a mechanism to accommodate them without special tokens. 
We attribute this to the fact that, unlike discriminative ViTs, all patch tokens in DiTs contribute to the loss, leaving no room for outliers.
When registers are introduced, they do not participate in the loss, thereby providing convenient slots for high-norm outliers.

In App.~\ref{app:t2i}, we find that patch tokens can occasionally become outliers in large text-to-image DiTs, but this behavior is inconsistent and vanishes in the later layers. This further suggests the need for dedicated outlier slots, since patches are constrained by the loss and cannot sustain high norms.

\subsection{Register Tokens Lead to Cleaner Internal Feature Maps}
\label{sec:feature_map_anal}

\begin{figure*}[t!]
    \centering
\includegraphics[width=1.0\textwidth]{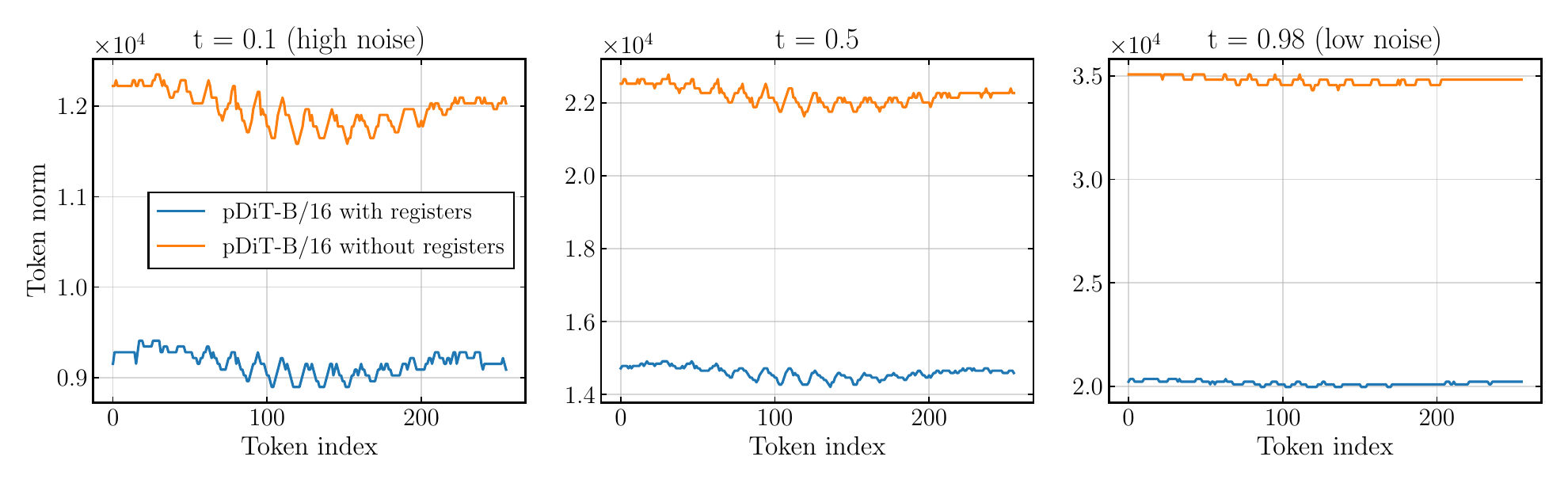}
    \caption{
        \textbf{Register tokens consistently reduce feature norms across patch tokens.} 
        We measure feature norms for patch tokens only, excluding registers, at different diffusion timesteps.
    }
\label{fig:analysis_jitb16_norms}
\end{figure*}

\begin{figure*}[t!]
    \centering
\includegraphics[width=1.0\textwidth]{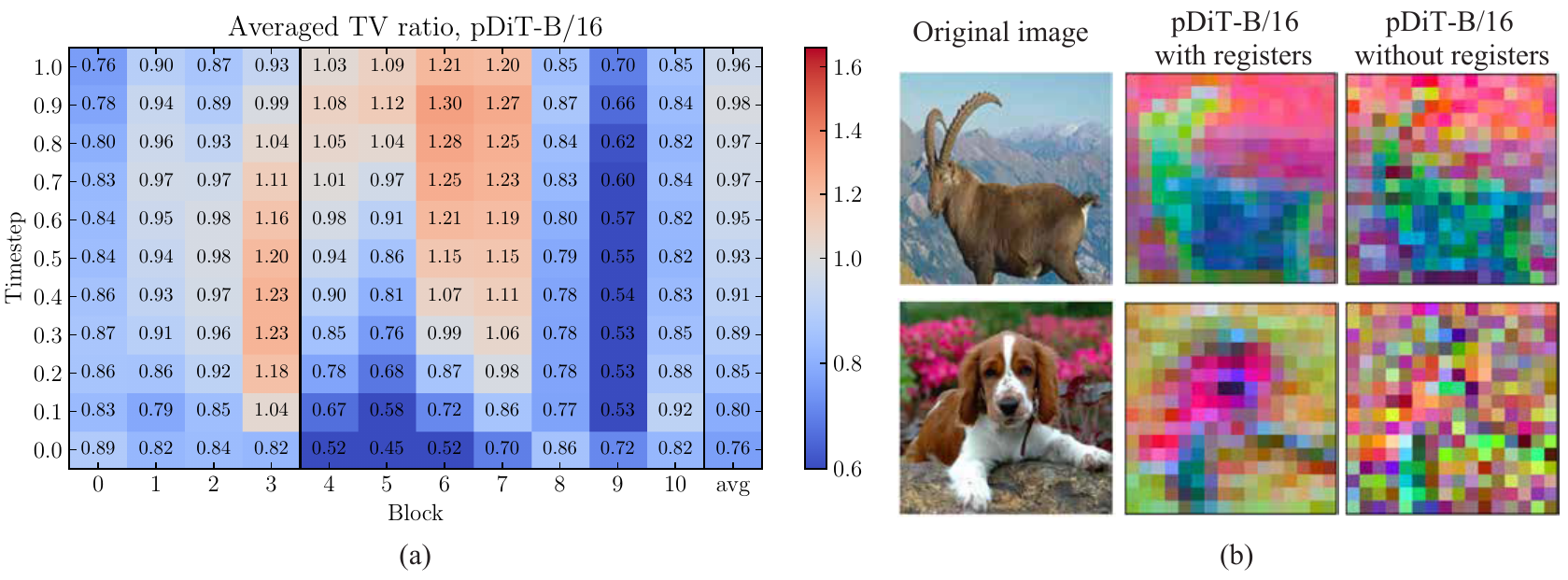}
    \caption{
        \textbf{Registers improve intermediate representations}.
        \textbf{(a)} Total Variation of intermediate features for models with and without register tokens. 
        We report the ratio (with / without registers); lower values indicate smoother features.
        Registers improve feature smoothness most at high noise levels ($t \in [0, 0.2]$).
        \textbf{(b)} PCA visualizations of feature maps at $t{=}0.1$ qualitatively confirm this effect.
    }
\label{fig:analysis_tv_loss_ratio}
\end{figure*}

The previous analysis shows that register tokens significantly improve pDiT performance, but their functional role remains unclear. 
In particular, their effect differs from that in ViTs, where registers primarily absorb pre-existing outliers. 
We therefore investigate how registers influence the internal representations of diffusion models, focusing on pixel-space models where their impact is strongest.

First, we find that register tokens influence all image tokens by
consistently reducing their feature norms (Figure~\ref{fig:analysis_jitb16_norms}). 
Interestingly, DINOv2 register tokens do not exhibit this behavior for non-outlier patch tokens (Figure~\ref{fig:app_analysis_dino_norms}). 

A possible explanation is that larger feature norms may reflect high local variability in hidden representations.
In DiTs, such variability may arise from the requirement to carefully predict high-dimensional targets, causing all information, including global semantics and low-level signals, to propagate through patch tokens.
Register tokens may absorb part of this information, reducing patch-token norms and allowing patch tokens to form smoother, more spatially structured representations.

To examine this, we consider Total Variation (TV)~\cite{rudin1992nonlinear, aly2005image}, which measures spatial smoothness by quantifying intensity differences between adjacent pixels. 
In our case, we use TV to quantify the spatial smoothness of intermediate transformer features. 
Specifically, we extract feature maps after each transformer block at different diffusion timesteps and compute their TV averaged over $1$K images.
We then evaluate TV value ratios (with registers / without registers).

We present the results in Figure~\ref{fig:analysis_tv_loss_ratio}(a) and observe that the ratios remain below 1 at lower timesteps (noisier images) and gradually approach 1 at higher timesteps (less noisy images). 
This suggests that register tokens improve feature smoothness primarily at high-noise levels ($t\in[0, 0.2]$).
Figure~\ref{fig:analysis_tv_loss_ratio}(b) provides qualitative support for this observation: PCA visualizations of intermediate features at $t{=}0.1$ show that models with registers produce smoother and more coherent feature representations.

Note that high-noise levels are particularly important for flow-matching models in high-dimensional spaces~\cite{esser2024scaling,bfl2025representation,yun2025no}, as they shape the main image content. Therefore, the improved representations induced by registers at these stages provide a plausible explanation for the observed quality gains.

\subsection{Registers Do Both: Encode Global Information and Act as Norm Sinks}
\label{sec:linprob_anal}

\begin{figure*}[t!]
    \centering
\includegraphics[width=1\textwidth]{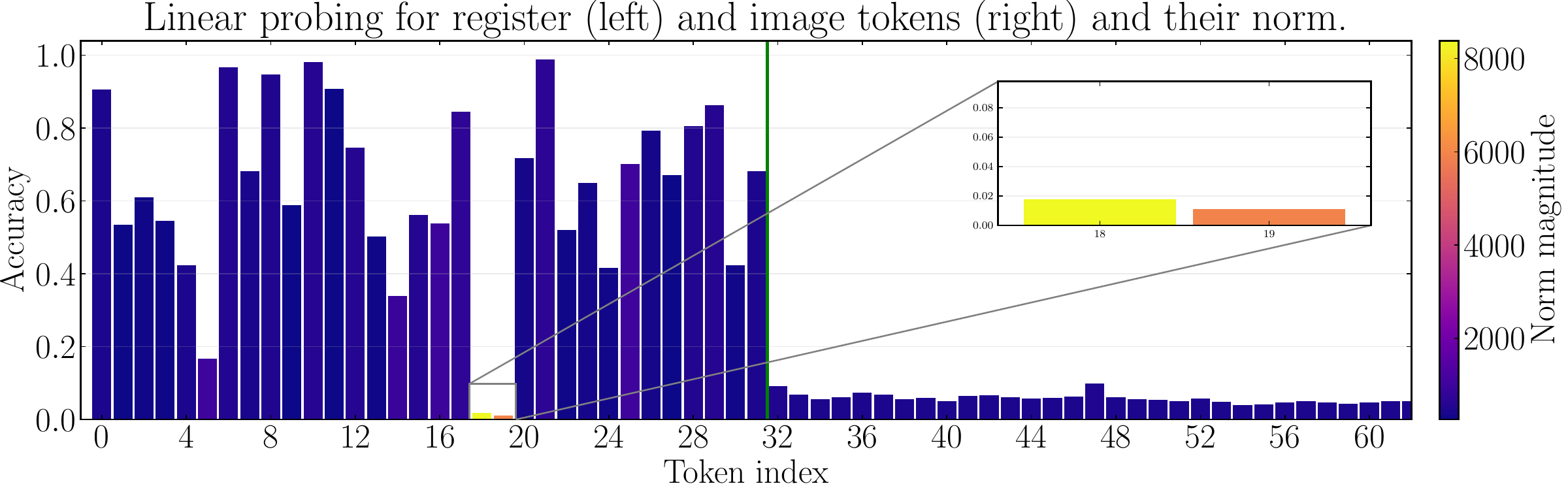}
\caption{
    \textbf{Register tokens act as both global information carriers and norm sinks}. 
    Linear probing reveals that low-norm register tokens encode meaningful global semantics and achieve strong classification accuracy, whereas low-accuracy registers exhibit much larger norms, suggesting that they primarily function as norm sinks that absorb magnitude from patch tokens.
}
\label{fig:analysis_probing_regs}
\end{figure*}

\begin{figure*}[t!]
    \centering
\includegraphics[width=1.0\textwidth]{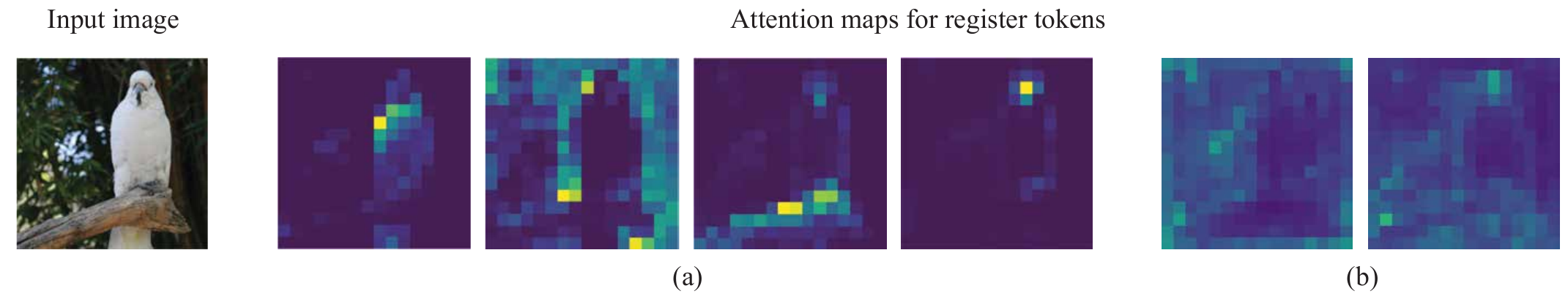}
    \caption{
    \textbf{(a) Registers with high probing accuracy encode diverse semantic information, whereas (b) low-accuracy norm sinks do not.}
    Visualizing register attention maps shows that some registers attend to distinct semantic regions, such as foreground objects and background areas. 
    In contrast, low-accuracy norm sinks do not show meaningful semantic structure.
}
\label{fig:analysis_attn_maps_regs}
\end{figure*}

Next, we find that, beyond acting as norm sinks for patch tokens,
register tokens can encode diverse global information about the input
image. Specifically, we perform linear probing using register-token features extracted from an intermediate transformer block (the 5th out of 12 blocks).

Figure~\ref{fig:analysis_probing_regs} shows that classification accuracy is highly diverse: some tokens achieve high accuracy ($\approx 0.9$), others moderate ($\approx 0.4$), and some very low ($\approx 0.02$). 
These results suggest the following: (a) tokens with the highest norms act as norm sinks, yielding the lowest accuracy; (b) tokens with moderate norms encode diverse information about the image beyond class-specific features.

To further validate that non-sink tokens encode diverse information, we visualize attention maps of different register tokens (Figure~\ref{fig:analysis_attn_maps_regs}a). 
We observe that registers attend to distinct semantic regions of the image. 
In the example, some tokens focus on background elements (e.g., jungle), while others attend to object parts (e.g., bird, branch, beak). 
In contrast, norm sinks do not encode meaningful semantic information (Figure~\ref{fig:analysis_attn_maps_regs}b).

\paragraph{Discussion.}
The insights from Sections~\ref{sec:feature_map_anal} and ~\ref{sec:linprob_anal} may relate to recent representation-alignment methods~\cite{yu2024representation, singh2025matters}, which improve DM convergence by aligning diffusion internal representations with vision encoders, e.g., DINOv2. 
Notably, iREPA~\cite{singh2025matters} shows that generation quality benefits more from aligning spatial structure than global semantics.
This aligns with our analysis: register tokens absorb global semantic information while improving the spatial coherence of patch tokens, suggesting that registers may play a regularizing role similar to REPA.
We therefore explore whether register tokens complement REPA-like objectives in App.~\ref{app:other}.

\begin{table*}[t]
\centering
\begin{minipage}[t]{0.49\textwidth}
\centering
\small
\setlength{\tabcolsep}{8pt}
\renewcommand{\arraystretch}{1.15}

\resizebox{\linewidth}{!}{%
\begin{tabular}{lccc}
\toprule[1.2pt]
& \textbf{RAE-space} & \textbf{VAE-space} & \textbf{Pixel-space} \\
& \textbf{DiT$^{\textbf{DH}}$ backbone} & \textbf{SiT backbone} & \textbf{pDiT backbone} \\
\midrule

\rowcolor{tabsection}
& \multicolumn{3}{c}{\textit{Base size}} \\
w/ reg.
& \cellcolor{tabred} 7.48
& \cellcolor{tabamber} \textbf{9.40}
& \cellcolor{tabgreen} \textbf{5.30} \\
w/o reg.
& \cellcolor{tabred} \textbf{6.58}
& \cellcolor{tabamber} 10.40
& \cellcolor{tabgreen} 7.39 \\
\midrule

\rowcolor{tabsection}
& \multicolumn{3}{c}{\textit{Large size}} \\
w/ reg.
& \cellcolor{tabred} 4.44
& \cellcolor{tabamber} \textbf{2.38}
& \cellcolor{tabgreen} \textbf{2.69} \\
w/o reg.
& \cellcolor{tabred} \textbf{3.91}
& \cellcolor{tabamber} 2.53
& \cellcolor{tabgreen} 3.52 \\
\bottomrule[1.2pt]
\end{tabular}
}
\vspace{2mm}
\caption{
    \textbf{Registers are more effective in pixel-space.}
    FID comparison with and without registers across training spaces and model sizes.
    Registers help pixel-space models most, moderately help VAE-space, and hurt RAE-space.
}
\label{tab:analysis_register_spaces}
\end{minipage}
\hfill
\begin{minipage}[t]{0.49\textwidth}
\centering
\small
\setlength{\tabcolsep}{6.8pt}

\resizebox{0.85\linewidth}{!}{%
\begin{tabular}{l c c c c c c}
\toprule[1.2pt]
 & \multicolumn{3}{c}{Registers configuration} & \multicolumn{3}{c}{FID at Epoch} \\
\cmidrule(lr){2-4} \cmidrule(lr){5-7}
 & Size & Start & End & 40 & 80 & 120 \\
\midrule

\multirow{6}{*}{w/ reg.}
& $32$ & $4$ & $11$ & ${37.7}$ & $\mathbf{9.59}$ & $\mathbf{6.45}$ \\
& $32$ & $4$ & $9$ & $\mathbf{36.9}$ & $9.95$ & $6.65$ \\
& $32$ & $0$ & $11$ & $59.7$ & $19.3$ & $11.9$ \\
& $32$ & $0$ & $4$  & $62.4$ & $19.6$ & $12.3$ \\
& $16$ & $4$ & $11$ & $40.4$ & $10.2$ & $6.80$ \\
& $4$  & $4$ & $11$ & $46.3$ & $12.8$ & $8.37$ \\

\midrule

w/o reg.
& $-$ & $-$ & $-$ & $60.6$ & $18.4$ & $11.1$ \\

\bottomrule[1.2pt]
\end{tabular}
}

\vspace{2mm}
\caption{
    \textbf{Registers are effective in deeper layers.} 
    Unlike DINOv2, pDiT-B/16 benefits from register tokens only when they are introduced after the first $4$ layers.
    Increasing the number of registers further improves performance.
}
\label{tab:analysis_register_start}
\end{minipage}
\end{table*}

\subsection{Registers Are More Effective in Pixel Space} \label{sec:pixel_space}

In Table~\ref{tab:analysis_register_spaces}, we compare the models operating in different spaces with and without registers.
We find that register tokens show the largest improvements in pixel space, provide smaller gains in VAE space~\cite{zheng2026diffusion}, and, interestingly, degrade performance in RAE-based models~\cite{zheng2026diffusion}.

To isolate the effect of different diffusion backbones, we apply the pDiT backbone to RAE and VAE spaces as well.
The results presented in Table~\ref{tab:app_analysis_register_spaces} show the same trend.
This indicates that the effect of register tokens is not related to corresponding architectural differences.

To better understand this effect, we analyze the smoothness of intermediate representations in DiTs without registers using TV. 
As shown in Figure~\ref{fig:app_analysis_tvs_models}, pixel-space models produce the least smooth (i.e., noisiest) intermediate features compared to latent DMs. 
In addition, we find that pixel-space models exhibit the highest feature norms for all patch tokens (Figure~\ref{fig:app_analysis_norms_models}), further supporting this observation.

These findings further suggest that training  DMs in pixel space is inherently more challenging and requires stronger regularization. In contrast, latent spaces are more structured and lower-dimensional, where imperceptible noise and fine-grained details are compressed. As a result, register tokens appear less critical for latent-space models, while serving as an effective mechanism for improving degraded representations in pixel-space DiTs.

\subsection{Registers Are Effective in Deeper Layers}\label{sec:deeper}

Next, we ablate both the number of register tokens and the transformer blocks in which they are introduced. We consider pDiT-B with $12$ layers. Based on the results in Table~\ref{tab:analysis_register_start}, we observe two key differences compared to standard ViTs. We explore additional configurations in Table~\ref{tab:app_analysis_register_start}.

First, pDiTs benefit from enabling register tokens only in deeper blocks ($4$--$11$), whereas ViTs use them from the first layer~\cite{darcet2023vision}.  
For example, applying registers throughout all layers ($0$--$11$) provides performance comparable to the model without registers. 
A similar effect appears for early-layer registers ($0$--$4$). Moreover, the $4$--$9$ configuration suggests that the final layers contribute less.

Previously, we found that register tokens encode diverse semantic information about the input image and help form more structured representations.
We hypothesize that their ineffectiveness in early layers stems from the lack of semantic structure at this stage. 
Specifically, early-layer registers primarily capture low-level or non-informative signals, providing weak conditioning to subsequent layers and ultimately degrading performance.
To support this observation, Figure~\ref{fig:app_analysis_probing} compares linear probing results for models with register tokens introduced at layers $4$ and $0$. When registers are enabled from the beginning, many non-sink registers capture little semantic information after the $5$th block, resulting in weak signals that are subsequently propagated to further layers.

Second, we observe that pDiTs require more registers than ViTs. While $4$ registers are typically sufficient for ViTs~\cite{darcet2023vision, simeoni2025dinov3}, pDiT-B achieves the best performance with $32$ tokens. 

\subsection{Registers Are Implicitly Present in Existing Diffusion Transformers}\label{sec:incontext}

\begin{figure*}[t!]
    \centering
\includegraphics[width=1.0\textwidth]{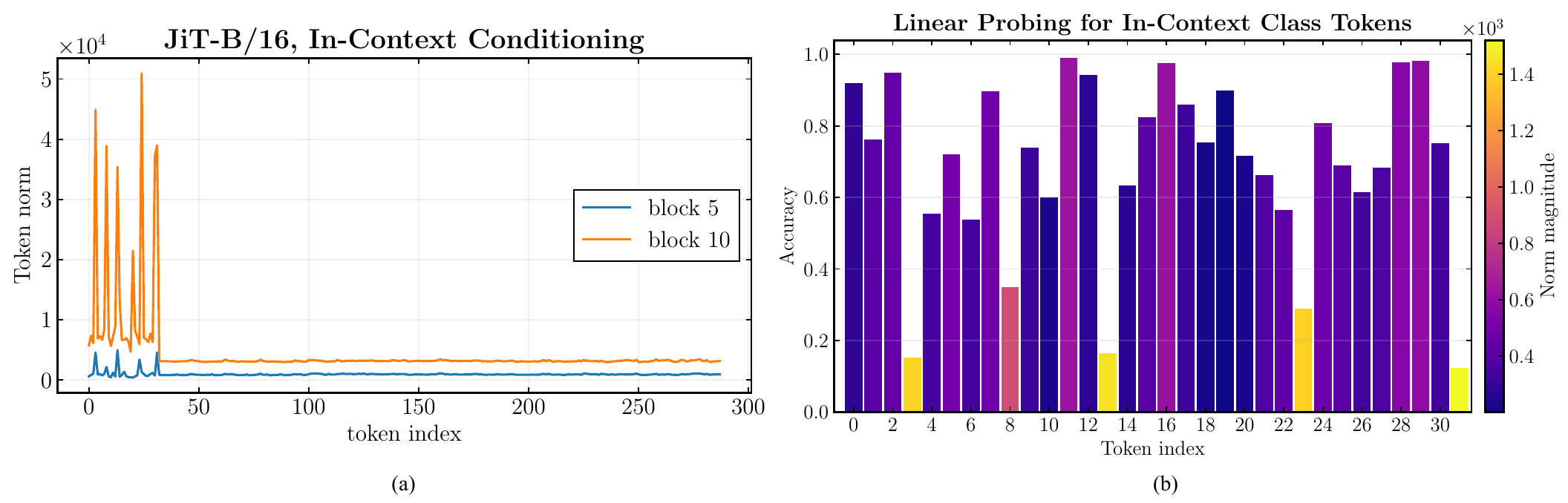}
    \caption{\textbf{In-context class tokens act as registers.} (a) Certain tokens acquire disproportionately high feature norms, functioning as norm sinks. (b) Some tokens encode broad global information, rather than purely class-specific features as originally intended.}
\label{fig:analysis_incontext_norms_probing}
\end{figure*}

Recent DiTs incorporate conditioning signals (e.g., text or class labels) by appending additional tokens to the sequence of image patches and processing them jointly through shared transformer layers. 
For example, JiT~\cite{li2025back}, a pixel-space DiT, employs in-context conditioning by adding duplicated class embeddings to the input sequence, leading to notable improvements in generation quality.
This raises the question of whether such in-context tokens implicitly function as register tokens.

To address this, we train JiT-B/16 with in-context conditioning, using the same diffusion backbone and number of tokens as in the register setting. 
Thus, the only difference between JiT-B/16 and pDiT-B/16 lies in the additional token sequence (in-context vs register tokens).

Then, we measure the norms of in-context tokens (Figure~\ref{fig:analysis_incontext_norms_probing}a) and evaluate their representations using linear probing (Figure~\ref{fig:analysis_incontext_norms_probing}b). 
Interestingly, we observe similar behavior as for registers: (a) some tokens encode diverse global information rather than only class-specific features as originally intended, 
(b) while others act as norm sinks. 
This suggests that {in-context tokens implicitly behave as registers}.

Moreover, Table~\ref{tab:analysis_jit_registers} compares models with registers, without registers, and with in-context conditioning. We find that most of the improvement over the baseline comes from the presence of pure register tokens rather than from the additional class information itself, helping explain the large quality gains from in-context conditioning.
However, in-context conditioning further improves performance, suggesting that such tokens help the model form better initial representations. 

We note that JiT~\cite{li2025back} introduces in-context tokens only in deeper layers, which motivates our ablation study on the register starting layer in Section~\ref{sec:deeper}.

We also analyze large-scale text-to-image models based on MM-DiT~\cite{esser2024scaling}, where text tokens are appended to image tokens. Interestingly, we observe a similar phenomenon: some text tokens exhibit high-norm outlier behavior (Figure~\ref{fig:app_analysis_t2i}), suggesting they may act as implicit registers.
\section{Register Guidance}

\begin{figure*}[t!]
    \centering
\includegraphics[width=1\textwidth]{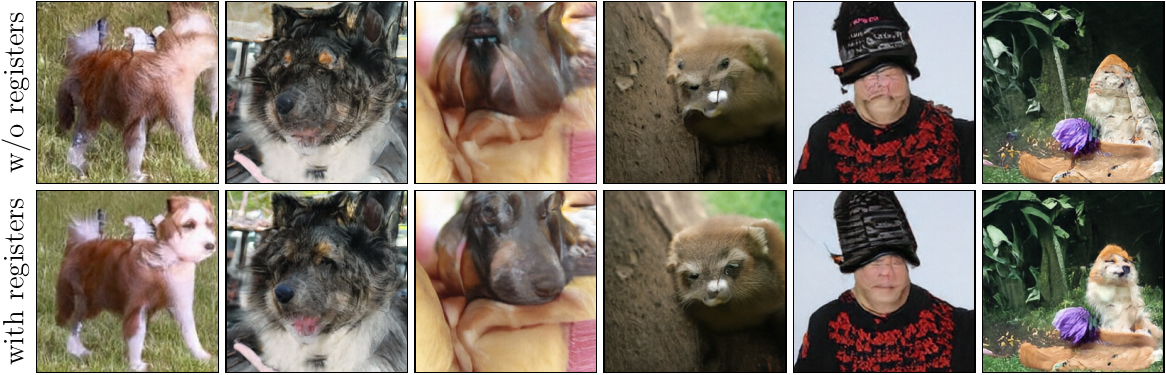}
\caption{
    \textbf{Registers improve image quality while preserving the overall image content.}
    We compare unconditional generations from the JiT-B/16 model without register tokens (top) and with register tokens (bottom), isolating the effect of registers from class conditioning.
    Using registers produces more coherent object structure and cleaner visual details across samples.
}
\label{fig:method_uncond_with_without_regs}
\end{figure*}

\begin{figure*}[t!]
    \centering
\includegraphics[width=1\textwidth]{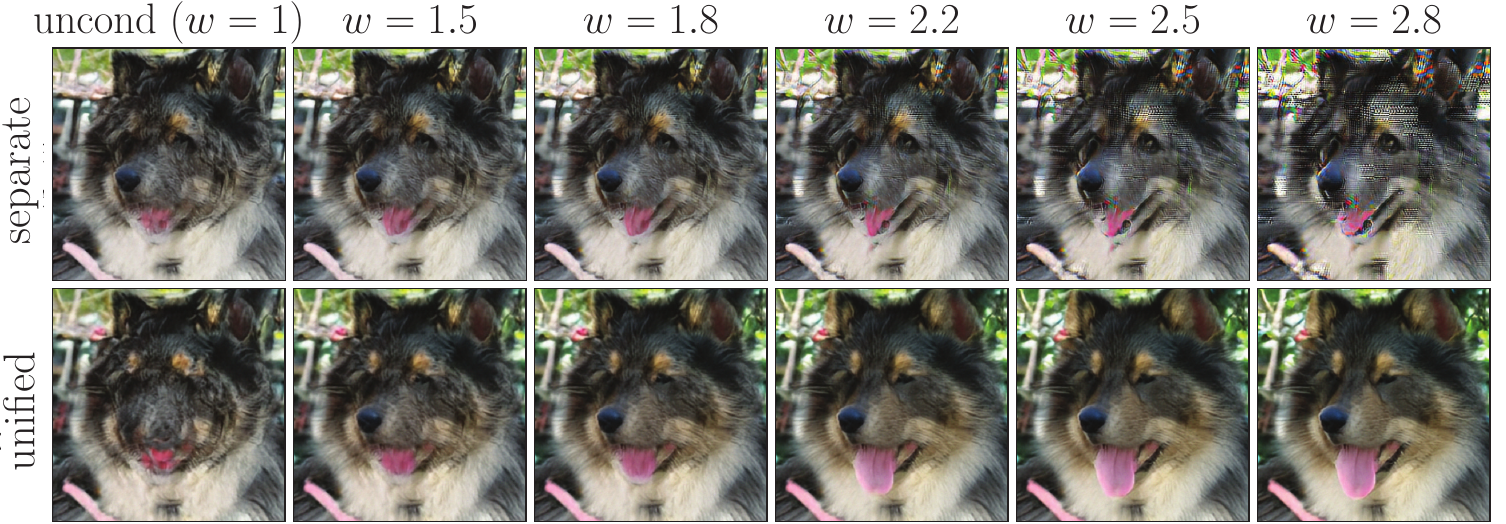}
\caption{
    \textbf{The influence of registers on improving image quality can be enhanced.}
    We apply Register Guidance using predictions with and without registers for different guidance scales $w$.
    Using two separate models introduces artifacts at large $w$ (top), whereas the unified model preserves the overall content and improves structure and details without introducing artifacts (bottom).
}
\label{fig:method_uncond_guidance}
\end{figure*}

\newlength{\snippetheight}
\setlength{\snippetheight}{52mm} 

In our method, register tokens are initialized from copied class embeddings, corresponding to in-context conditioning~\cite{li2025back, lu2026one}, which gives slightly better performance. We still refer to them as registers because both variants serve a similar function. Thus, our approach builds on JiT models~\cite{li2025back}.

{\textbf{Intuition.} Based on our analysis, registers improve generation quality, functioning as tokens that enhance internal representations and encode global information about the input image. So, we first investigate how generations specifically improve when registers are included.}

To this end, we compare generations from a JiT-B/16 model with and without registers.  Importantly, we perform this comparison in the unconditional setting to remove the influence of class information, which is encoded in the registers. This allows us to isolate the effect of registers themselves and examine how they improve generation quality beyond acting as an additional conditioning signal.

Figure~\ref{fig:method_uncond_with_without_regs} shows a clear effect of register tokens. With registers, generations have more coherent object structures and cleaner visual details, while the overall image content remains preserved.

These results motivate the following question: \textit{can the influence of registers be enhanced to improve quality even further?} We draw inspiration from AutoGuidance~\cite{karras2024guiding}, which improves sample quality by guiding a model with a weaker version of itself rather than with an unconditional model. Our setting follows a similar intuition: the same model without registers can be viewed as a weaker counterpart to the model with registers. This connection is possible because both variants preserve similar image content, while the version without registers exhibits stronger artifacts and less coherent structure. Thus, their difference provides a meaningful direction for improving generation quality.

\begin{table}[t!]
\centering

\lstset{
  language=trainalgo,
  basicstyle=\ttfamily\fontsize{7.5}{8.5}\selectfont,
  showstringspaces=false,
  frame=none,
  xleftmargin=0pt,
  xrightmargin=0pt,
  aboveskip=0pt,
  belowskip=0pt,
  keepspaces=true,
  columns=fullflexible,
  breaklines=true
}

\begin{minipage}[t][\snippetheight][t]{0.315\linewidth}
\hrule height 0.8pt
\vspace{0.25em}

{\snippettitle{snip:rg_inference}{RG, Inference}}

\vspace{0.25em}
\hrule height 0.5pt
\vspace{0.35em}

\begin{lstlisting}
# z: current samples at t
# y: class label
# w_rg: register guidance scale

x_r  = net(z, t, y, regs=True)
x_nr = net(z, t, y, regs=False)

v_r  = (x_r  - z) / (1 - t)
v_nr = (x_nr - z) / (1 - t)

v_rg = v_nr + w_rg * (v_r - v_nr)

z_next = z + (t_next - t) * v_rg
\end{lstlisting}

\vfill
\hrule height 0.5pt
\end{minipage}
\hfill
\begin{minipage}[t][\snippetheight][t]{0.315\linewidth}
\hrule height 0.8pt
\vspace{0.25em}

{\snippettitle{snip:rg_cfg_inference}{RG + CFG, Inference}}

\vspace{0.25em}
\hrule height 0.5pt
\vspace{0.35em}

\begin{lstlisting}
# y0: null label
# w_cfg: CFG scale

x_r  = net(z, t, y,  regs=True)
x_nr = net(z, t, y,  regs=False)
x_u  = net(z, t, y0, regs=True)

v_r  = (x_r  - z) / (1 - t)
v_nr = (x_nr - z) / (1 - t)
v_u  = (x_u  - z) / (1 - t)

v = v_u + w_cfg * (v_r - v_u) 
        + w_rg * (v_r - v_nr) 

z_next = z + (t_next - t) * v
\end{lstlisting}

\vfill
\hrule height 0.5pt
\end{minipage}
\hfill
\begin{minipage}[t][\snippetheight][t]{0.315\linewidth}
\hrule height 0.8pt
\vspace{0.25em}

{\snippettitle{snip:rg_train}{RG, Training}}

\vspace{0.25em}
\hrule height 0.5pt
\vspace{0.35em}

\begin{lstlisting}
# net(z, t, regs): network
# x: training batch
# p: prob. of dropping registers

t, e = sample_t(), randn_like(x)
z = t * x + (1 - t) * e
v = (x - z) / (1 - t)

# Drop registers during training
use_regs = rand() >= p

x_pred = net(z, t, regs=use_regs)
v_pred = (x_pred - z) / (1 - t)

loss = l2_loss(v - v_pred)
\end{lstlisting}

\vfill
\hrule height 0.5pt
\end{minipage}
\hfill

\end{table}

\textbf{Our method.} We call our approach \textit{Register Guidance (RG)}. It is simple to implement and apply, as described in Snippets, \ref{snip:rg_inference}, \ref{snip:rg_cfg_inference} and \ref{snip:rg_train}.

\textit{Inference.}{ First, we describe the inference procedure in Snippet~\ref{snip:rg_inference}. RG follows a similar principle to CFG, however, instead of using an unconditional prediction, it uses the prediction of the model evaluated without registers. We introduce an additional hyperparameter \texttt{\textcolor{MidnightBlue}{w\_rg}} to control the RG strength.}

{Moreover, in Snippet~\ref{snip:rg_cfg_inference}, we show how to combine RG with CFG. Since registers carry a source of information beyond the class label, RG can be combined with CFG in a complementary way. This strategy further improves generation quality.}

\textit{Training.} We apply RG (Snippet \ref{snip:rg_inference}) in the unconditional setting using two separate models: JiT-B/16 models trained with and without registers. Figure~\ref{fig:method_uncond_guidance} (top) presents the results. Interestingly, we observe quality improvements, but also the introduction of artifacts at larger guidance scales. This suggests that the model difference is meaningful, as it improves quality, but also imperfectly aligned, as reflected by the emerging artifacts. More examples are shown in Figure~\ref{fig:app_uncond_guidance_separate_full}.

The artifacts arising from using two separate models suggest that better alignment between predictions is needed. This motivates us to train a single model that supports generation both with and without registers, improving alignment between the two predictions. We describe the training procedure in Snippet~\ref{snip:rg_train}. We use the original training procedure~\cite{li2025back}, but occasionally drop all register tokens from the sequence, enabling the model to operate both with and without registers.
We introduce an additional hyperparameter \texttt{\textcolor{MidnightBlue}{p}}, which controls the probability of dropping registers.

Figure~\ref{fig:method_uncond_guidance} (bottom) shows RG in the unconditional setting using a single model. The artifacts disappear, indicating better alignment between the two predictions, while generation quality continues to improve. 

{Importantly, during training, we use small values of \texttt{\textcolor{MidnightBlue}{p}}, which preserve the performance of the original register-based model while enabling the model to operate without registers. Technically, we use separate batches for the two regimes, which does not degrade performance. Although the same behavior could be achieved by masking register tokens in self-attention.

\textit{Discussion}. {Our approach can be understood as a specific instance of AutoGuidance~\cite{karras2024guiding}. However, our method gives a concrete way to obtain a weaker model, while in AG this choice can be challenging because it requires careful manual design~\cite{zhou2026guiding}. Moreover, we find that AG does not improve performance in the JiT setup, which is further confirmed by~\cite{baade2026latent}.}

\begin{figure*}[t!]
    \centering
\includegraphics[width=1\textwidth]{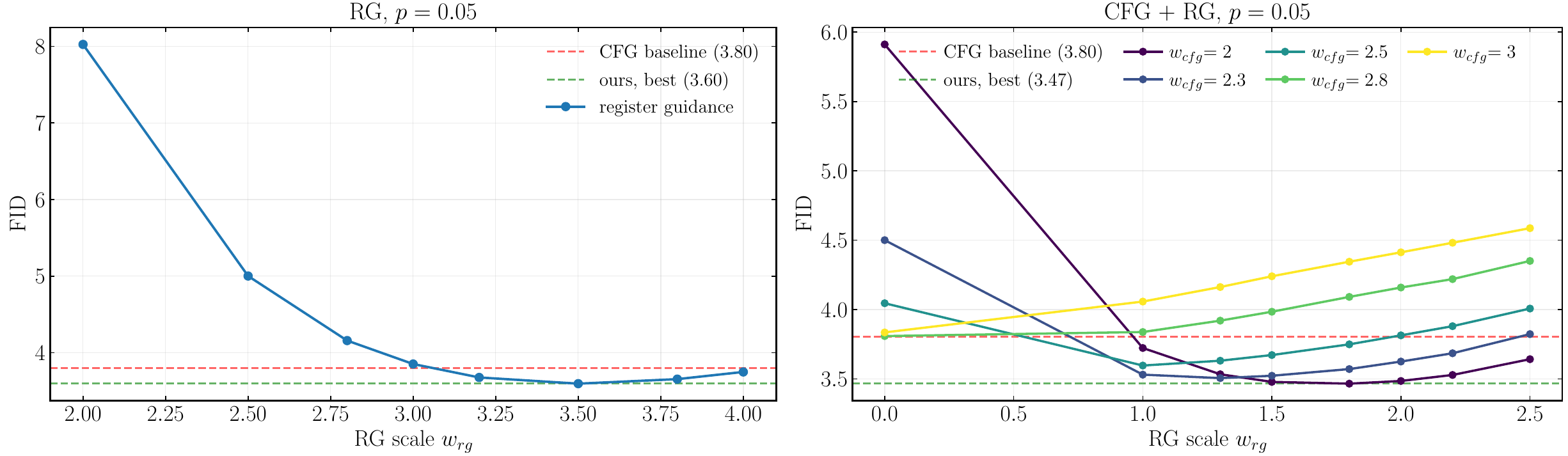}
\caption{
    \textbf{Effect of \texttt{\textcolor{MidnightBlue}{w\_rg}} under fixed \texttt{\textcolor{MidnightBlue}{p=0.05}}.}
    Left: RG improves FID as the scale increases, with the best result at \texttt{\textcolor{MidnightBlue}{w\_rg=3.5}}.
    Right: RG remains complementary to CFG, further improving performance over the CFG baseline, with the best setting at \texttt{\textcolor{MidnightBlue}{w\_rg=1.8}} and \texttt{{w\_cfg=2.0}}.
}
\label{fig:exps_rg_cfg}
\end{figure*}

\begin{table*}[t!]
\centering

\begin{minipage}[t]{0.49\textwidth}
\centering
\small
\setlength{\tabcolsep}{3pt}
\renewcommand{\arraystretch}{1.12}

\begin{tabular*}{\linewidth}{@{\extracolsep{\fill}}lccccc@{}}
\toprule[1.2pt]
$p$
& $0.0$ & $0.1$ & $0.05$ & $0.03$ & $0.01$ \\
\midrule
CFG
& $3.71$ & $3.96$ & $3.80$ & $\mathbf{3.69}$ & $3.86$ \\

RG
& $13.13$ & $4.55$ & $3.60$ & $\mathbf{3.46}$ & $3.77$ \\

RG+CFG
& $8.02$ & $3.85$ & $3.47$ & $\mathbf{3.32}$ & $3.64$ \\
\bottomrule[1.2pt]
\end{tabular*}

\vspace{2mm}
\caption{
\textbf{Effect of \texttt{\textcolor{MidnightBlue}{p}} under fixed \texttt{\textcolor{MidnightBlue}{w\_rg}}.}
We vary \texttt{\textcolor{MidnightBlue}{p}} during training, where \texttt{\textcolor{MidnightBlue}{p=0}} corresponds to the original setting~\cite{li2025back}. 
\texttt{\textcolor{MidnightBlue}{p=0.03}} gives the best performance without degrading the CFG baseline.
}
\label{tab:exps_rg_p_ablation}
\end{minipage}
\hfill
\begin{minipage}[t]{0.49\textwidth}
\centering
\small
\setlength{\tabcolsep}{3pt}
\renewcommand{\arraystretch}{1.12}

\begin{tabular*}{\linewidth}{@{\extracolsep{\fill}}lcccccc@{}}
\toprule[1.2pt]
& B/16 & L/16 & B/32 & L/32 \\
\midrule
CFG
& $3.71$ & $2.47$ & $4.12$ & $2.69$ \\

RG
& $3.46$ & $2.18$ & $4.01$ & $2.47$ \\

RG+CFG
& $\mathbf{3.32}$ & $\mathbf{2.16}$ & $\mathbf{3.69}$ & $\mathbf{2.41}$ \\
\bottomrule[1.2pt]
\end{tabular*}

\vspace{2mm}
\caption{
    \textbf{Register Guidance improves across model scales and image sizes.}
    With the best \texttt{\textcolor{MidnightBlue}{p}} and \texttt{\textcolor{MidnightBlue}{w\_rg}}, RG improves over CFG, and RG+CFG gives the best FID in all settings.
}
\label{tab:exps_rg_scale_ablation}
\end{minipage}

\end{table*}

\definecolor{tabsection}{HTML}{E6E6E6}

\begin{table*}[t!]
\centering

\begin{minipage}[t]{0.49\textwidth}
\vspace{0pt}
\centering
\small
\setlength{\tabcolsep}{4pt}
\renewcommand{\arraystretch}{1.05}

\resizebox{\linewidth}{!}{%
\begin{tabular}{@{}lccc@{}}
\toprule[1.2pt]
\textbf{ImageNet $256{\times}256$} & \textbf{Params} & \textbf{FID$\downarrow$} & \textbf{IS$\uparrow$} \\
\midrule

\rowcolor{tabsection}
\multicolumn{4}{@{}l}{\textit{Latent-space Diffusion}} \\
DiT~\cite{peebles2023scalable}, XL/2              & $675$M  & $2.27$ & $278.2$ \\
SiT~\cite{ma2024sit}, XL/2              & $675$M  & $2.06$ & $277.5$ \\
REPA~\cite{yu2024representation}, SiT-XL/2        & $675$M  & $1.42$ & $305.7$ \\
RAE~\cite{zheng2026diffusion}, DiT$^{\mathrm{DH}}$-XL/2
                             & $839$M & $\mathbf{1.13}$ & $262.6$ \\

\midrule
\rowcolor{tabsection}
\multicolumn{4}{@{}l}{\textit{Pixel-space Diffusion}} \\
VDM++~\cite{kingma2023understanding}, UViT/2              & $2$B   & $2.12$ & $267.7$ \\
SiD2~\cite{hoogeboom2025simpler}, UViT/1          & N/A    & $\mathbf{1.38}$ & -- \\
PixelFlow~\cite{chen2025pixelflow}, XL/4        & $677$M & $1.98$ & $282.1$ \\
PixNerd~\cite{wang2025pixnerd}, XL/16        & $700$M & $2.15$ & $297.0$ \\
{JiT~\cite{li2025back}, B/16}         & $131$M & $3.66$ & $275.1$ \\
{JiT~\cite{li2025back}, L/16}          & $459$M & $2.36$ & $298.5$ \\
{JiT~\cite{li2025back}, H/16}           & $953$M & $1.86$ & $303.4$ \\
{JiT~\cite{li2025back}, G/16}          & $2$B   & $1.82$ & $292.6$ \\
\midrule
\textbf{JiT-B/16, CFG + RG}        & $131$M & $2.96$ & $251.1$ \\
\textbf{JiT-L/16, CFG + RG}         & $459$M & $2.00$ & $274.5$ \\
\textbf{JiT-H/16, CFG + RG}        & $953$M & $1.80$ & $270.1$ \\
\bottomrule[1.2pt]
\end{tabular}
}

\vspace{2mm}
\caption{
   \textbf{ImageNet $256{\times}256$ results.}
}
\label{tab:exps_imagenet256_comparison}
\end{minipage}
\hfill
\begin{minipage}[t]{0.49\textwidth}
\vspace{0pt}
\centering
\small
\setlength{\tabcolsep}{4pt}
\renewcommand{\arraystretch}{1.114}

\resizebox{\linewidth}{!}{%
\begin{tabular}{@{}lccc@{}}
\toprule[1.2pt]
\textbf{ImageNet $512{\times}512$} & \textbf{Params} & \textbf{FID$\downarrow$} & \textbf{IS$\uparrow$} \\
\midrule

\rowcolor{tabsection}
\multicolumn{4}{@{}l}{\textit{Latent-space Diffusion}} \\
DiT~\cite{peebles2023scalable}, XL/2            & $675$M  & $3.04$ & $240.8$ \\
SiT~\cite{ma2024sit}, XL/2             & $675$M  & $2.62$ & $252.2$ \\
REPA~\cite{yu2024representation}, SiT-XL/2        & $675$M  & $2.08$ & $274.6$ \\
RAE~\cite{zheng2026diffusion}, DiT$^{\mathrm{DH}}$-XL/2
                             & $839$M & $\mathbf{1.13}$ & $259.6$ \\

\midrule
\rowcolor{tabsection}
\multicolumn{4}{@{}l}{\textit{Pixel-space Diffusion}} \\
VDM++~\cite{kingma2023understanding}, UViT/4              & $2$B   & $2.65$ & $278.1$ \\
SiD2~\cite{hoogeboom2025simpler}, UViT/2          & N/A    & $\mathbf{1.48}$ & -- \\
PixNerd~\cite{wang2025pixnerd}, XL/16        & $700$M & $2.84$ & $245.6$ \\
{JiT~\cite{li2025back}, B/32}         & $133$M & $4.02$ & $271.0$ \\
{JiT~\cite{li2025back}, L/32}           & $462$M & $2.53$ & $299.9$ \\
{JiT~\cite{li2025back}, H/32}         & $956$M & $1.94$ & $309.1$ \\
{JiT~\cite{li2025back}, G/32}          & $2$B   & $1.78$ & $306.8$ \\
\midrule
\textbf{JiT-B/32, CFG + RG}        & $133$M & $3.34$ & $255.9$ \\
\textbf{JiT-L/32, CFG + RG}         & $462$M & $2.23$ & $283.5$ \\
\textbf{JiT-H/32, CFG + RG}        & $956$M & $1.87$ & $286.9$ \\
\bottomrule[1.2pt]
\end{tabular}
}

\vspace{2mm}
\caption{
   \textbf{ImageNet $512{\times}512$ results.}
}
\label{tab:exps_imagenet512_comparison}
\end{minipage}
\end{table*}

\textbf{Results.} We test our approach on ImageNet 256 and 512 in the conditional setting. First, we ablate the influence of \texttt{\textcolor{MidnightBlue}{w\_rg}} and \texttt{\textcolor{MidnightBlue}{p}}. We use JiT-B/16 and report results for different hyperparameter values in Figure~\ref{fig:exps_rg_cfg} and Table~\ref{tab:exps_rg_p_ablation}. Under optimal parameters, RG improves performance across different model scales and image sizes (Table~\ref{tab:exps_rg_scale_ablation}). More details on hyperparameter selection are in App.~\ref{app:rg_details}.

Tables~\ref{tab:exps_imagenet256_comparison} and~\ref{tab:exps_imagenet512_comparison} compare our method with previous works, where our approach shows strong performance.
To achieve the best results, we use the original JiT~\cite{li2025back} checkpoints and fine-tune them with our training procedure. More experimental results are provided in App.~\ref{app:rg_results}.

\section{Discussion}

Registers in image DiTs represent a promising research direction, offering both analytical insights and practical implications. While this work has explored certain aspects of this topic, we believe considerable scope remains for further investigation.

One potential direction follows from the clearly distinct behaviors of register and patch tokens: developing dual-stream DiT architectures~\cite{esser2024scaling} that process them more effectively. We preliminarily explored several efficient dual-stream designs in App.~\ref{app:dual} but did not observe major improvements. Nevertheless, we believe this direction merits further exploration.

Moreover, our exploration of Register Guidance leaves room for further study. While we focus on pixel-space models in the class-conditional setting, extending Register Guidance to text-to-image models presents an interesting direction for future research~\cite{gan2025massive}.

Overall, we hope that our findings provide a useful step toward understanding registers in DiTs and encourage future work on methods that better exploit their potential.






\bibliography{neurips_2026}
\bibliographystyle{unsrt}


\newpage
\appendix
\section{Related Work}
\label{app:related}

\textbf{Attention Sinks in Large Language Models.} In autoregressive LLMs, attention sinks are a well-explored area \cite{xiao2023efficient, sun2024massive, gu2024attention, yona2025interpreting, qiu2025gated}. 
\cite{xiao2023efficient} first analyzes anomalies in attention and finds that a large portion of the attention score is allocated to the initial tokens, making them attention sinks. 
The authors propose preserving these initial register tokens during inference, which leads to a significant quality boost. \cite{sun2024massive} continues the sink analysis and finds that outliers appear in a few fixed feature dimensions, regardless of the input, as well as in two types of special tokens. 
\cite{gu2024attention} analyzes why and how attention sinks emerge during training. 
\cite{yona2025interpreting} explores the functional role of attention sinks, showing their connection to the repeated token divergence phenomenon. 
\cite{qiu2025gated} demonstrates that sparse gating can eliminate attention sinks.

\textbf{Attention Sinks in Diffusion Language Models.} 
\cite{rulli2025attention} extends the study of attention sinks to DLMs, showing that attention sinks persist, but with different behavior: in DLMs, the positions of attention sinks tend to shift during the generation process as tokens are progressively unmasked. 
Moreover, these sinks can be masked without significant degradation. 
\cite{zhang2026one} addresses this moving-sink behavior by adding an additional sink token that is globally visible to all other tokens while attending only to itself.
Such a token is shown to stabilize DLM inference.

\textbf{Attention Sinks in Video Diffusion Transformers.} 
\cite{wen2025analysis} presents the first analysis of attention sinks in video DiTs, highlighting similarities and differences with LLMs.
Notably, they find that these sinks are concentrated in the first frame, analogous to the <BOS> token in LLMs. 
\cite{liu2025rolling, shin2025motionstream} consider an autoregressive approach to video generation, predicting multiple frames simultaneously. 
Similar to LLMs~\cite{xiao2023efficient}, they find that it is important to retain the first frame, which acts as an attention sink. 
\cite{yi2025deep} extends the idea of keeping the initial frame by proposing Deep Sink, which aims to stabilize global context during long rollouts. 
Note that video DiTs use attention sinks mainly to enable long-horizon AR sampling, rather than to study their effect on denoising performance.

\textbf{Outliers in Feature Dimensions.} Another line of work studies outliers that emerge not across tokens, but within feature (channel) dimensions of learned representations~\cite{an2025systematic, simeoni2025dinov3, gan2026unleashing, gan2025massive}. 
Initially studied in LLMs~\cite{an2025systematic}, this phenomenon was later analyzed in DiTs~\cite{gan2026unleashing, gan2025massive}. 
In particular, \cite{gan2026unleashing} links massive activations to layer normalization, while \cite{gan2025massive} shows that manipulating them can improve fine-grained image details. 

Importantly, concurrent work by \cite{gan2025massive} is closely related to our Register Guidance, but uses a model with channel outliers removed as the weak model. They show that this mainly improves fine-grained details. In contrast, our approach uses a model without register tokens as the weak model, leading to substantially larger changes that improve not only fine details but also global structure and visual coherence. We hypothesize that removing register tokens produces a stronger weak counterpart than removing only a subset of channels, resulting in a more informative guidance signal.

\textbf{Register Tokens in Vision Transformers.}
Several works study the role and behavior of register tokens in ViTs~\cite{darcet2023vision, jiang2025vision, lappe2025register, shi2026vision, chen2025vision, wang2024sinder, marouani2026revisiting}. 
\cite{darcet2023vision} introduces register tokens to avoid sink artifacts in attention maps, improving ViT internal representations.
Subsequent studies further analyze their functional role, showing that they can influence feature aggregation~\cite{lappe2025register}. 
\cite{jiang2025vision, chen2025vision} study post-hoc or self-distilled ways to add registers to pretrained ViTs without full retraining. 
\cite{shi2026vision} argue that register tokens alone do not fully explain or resolve all ViT artifacts.
Recent large-scale ViTs also retain explicit outlier-handling mechanisms: DINOv3 adopts register tokens after comparing them with attention-bias and value-gating alternatives inspired by LLM outlier analyses~\cite{simeoni2025dinov3}.

\textbf{Attention Sinks in Text-to-Image DiTs.}
For text-to-image diffusion, \cite{yi2024towards} highlights the special role of text tokens across different denoising steps, while \cite{jamal2026diffusion} studies high-norm activations in pretrained text-to-image DiTs.
{Padding Tone~\cite{toker2025paddingtonemechanisticanalysis} further shows that auxiliary text tokens, such as padding tokens, can influence text-to-image generation. However, these works do not fully explain why such outliers can improve generation quality or what role they play within the model. Our work addresses this gap by systematically studying high-norm tokens and connecting them to register tokens, showing that registers play an important role in generation quality.}

To summarize, tokens with unusually large norms have been studied across different domains~\cite{su2026attention}. In this work, we extend this research direction to image diffusion transformers and show that they are particularly important for pixel-space models.

\section{Implementation Details}\label{app:implementation_details}

\subsection{Analysis Implementation Details}
In our implementation of pDiTs, we largely follow the JiT setup~\cite{li2025back}. The model uses flow matching with $x$-prediction and the forward process $ x_t = t x + (1-t)\epsilon$, where $t=1$ corresponds to clean data. We adopt the same diffusion backbone, but remove in-context conditioning and instead introduce register tokens implemented as trainable parameters without additional layers~\cite{darcet2023vision}.

We train models of three sizes $-$ B ($131$M), L ($459$M), and H ($953$M) $-$ on ImageNet $256\times256$ and ImageNet $512\times512$. For the B and L models, we use a batch size of $1024$, following JiT~\cite{li2025back}. Due to limited computational resources, the H model on ImageNet $512\times512$ is trained with a smaller batch size of $512$. All other training and inference settings follow JiT~\cite{li2025back}.

For PixelDiT~\cite{yu2025pixeldit}, we consider the XL model ($800$M) and train three models for $120$ epochs: with registers, without registers, and with in-context conditioning. We use the same configuration except for gradient clipping, where we use a smaller value of $0.5$ instead of $1.0$ from the original implementation for stability.

For the latent-space models presented in Table~\ref{tab:analysis_register_spaces}, we follow their original setups~\cite{zheng2026diffusion, ma2024sit}. For RAE-XL, we use a smaller batch size ($256$ instead of $1024$) due to computational constraints and train the model for $80$ epochs. For SiTs, we use the same batch size as in the original implementation and train the models for $300$ epochs. Register tokens are implemented in the same manner as in the pixel-space models.

\subsection{Register Guidance Implementation Details}
\label{app:rg_details}

\begin{table*}[t!]
\centering
\small
\renewcommand{\arraystretch}{1.35}
\setlength{\tabcolsep}{5pt}

\begin{tabular}{lcccccc}
\toprule
& & \multicolumn{2}{c}{RG} & \multicolumn{3}{c}{RG+CFG} \\
\cmidrule(r{8pt}){3-4}
\cmidrule(l{8pt}){5-7}
Model & CFG & RG & $p$ & RG & CFG & $p$ \\
\midrule
\addlinespace[2.5mm]

JiT-B/16
& \cfg{3.0}
& \rg{3.5}
& $0.03$
& \rgpart{1.8}
& \cfgpart{2.0}
& $0.03$ \\

\midrule
\addlinespace[1.5mm]

JiT-L/16
& \cfg{2.4}
& \rg{2.8}
& $0.03$
& \rgpart{1.4}
& \cfgpart{1.6}
& $0.03$ \\

\midrule
\addlinespace[1.5mm]

JiT-H/16
& \cfg{2.2}
& \rg{2.6}
& --
& \rgpart{1.1}
& \cfgpart{1.4}
& -- \\

\midrule
\addlinespace[1.5mm]

JiT-B/32
& \cfg{3.0}
& \rg{3.2}
& $0.03$
& \rgpart{1.8}
& \cfgpart{2.0}
& $0.03$ \\

\midrule
\addlinespace[1.5mm]

JiT-L/32
& \cfg{2.5}
& \rg{2.8}
& $0.03$
& \rgpart{1.4}
& \cfgpart{1.6}
& $0.03$ \\

\midrule
\addlinespace[1.5mm]

JiT-H/32
& \cfg{2.3}
& \rg{2.6}
& --
& \rgpart{1.1}
& \cfgpart{1.4}
& -- \\

\bottomrule
\end{tabular}

\caption{
\textbf{Training and inference configurations for CFG, RG, and RG+CFG.}
For RG and RG+CFG, we separately report the guidance parameters and register-drop probability $p$.
}
\label{tab:app_hyperparams}
\end{table*}

As discussed in the main text, we train a JiT model from scratch while periodically dropping registers with probability $p$. We compare three settings: CFG (baseline), RG, and RG+CFG. We train all models for $600$ epochs.
For all settings, we use the interval guidance strategy~\cite{kynkäänniemi2024applyingguidancelimitedinterval}. We present the optimal hyperparameters in Table~\ref{tab:app_hyperparams}.

Besides training from scratch, we also consider a fine-tuning setup (Tables~\ref{tab:exps_imagenet256_comparison} and~\ref{tab:exps_imagenet512_comparison}). In this case, we take pretrained JiT models~\cite{li2025back}, which cannot generate without registers, and fine-tune them for $50$ epochs using a lower learning rate of $2\mathrm{e}{-5}$. Moreover, we use a larger register-drop probability, $p=0.3$, while keeping the other parameters the same as in Table~\ref{tab:app_hyperparams}. This setup gives the best performance and shows that Register Guidance can be applied to already trained models with little additional effort.

\section{Additional Analysis Results}

\begin{table*}[t!]
\centering
\footnotesize
\renewcommand{\arraystretch}{1.12}

\newcommand{\noreg}{\shortstack{w/o regs.}}
\newcommand{\wreg}{\shortstack{w/ regs.}}

\setlength{\tabcolsep}{0pt}
\begin{tabular*}{\textwidth}{@{\extracolsep{\fill}}lccccccccc@{}}
\toprule[1.05pt]
& \multicolumn{3}{c}{pDiT-B/32}
& \multicolumn{3}{c}{pDiT-L/32}
& \multicolumn{3}{c}{pDiT-H/32} \\
\cmidrule(lr){2-4}
\cmidrule(lr){5-7}
\cmidrule(lr){8-10}
Epoch
& \noreg & \wreg & \fullgray{IC}
& \noreg & \wreg & \fullgray{IC}
& \noreg & \wreg & \fullgray{IC} \\
\midrule
$200$
& $8.67$ & $\mathbf{6.29}$ & \fullgray{$5.84$}
& $4.41$ & $\mathbf{3.33}$ & \fullgray{$3.28$}
& $3.47$ & $\mathbf{2.70}$ & \fullgray{$2.77$} \\

$600$
& $5.14$ & $\mathbf{4.20}$ & \fullgray{$4.12$}
& $3.12$ & $\mathbf{2.72}$ & \fullgray{$2.69$}
& $2.68$ & $\mathbf{2.08}$ & \fullgray{$2.08$} \\
\bottomrule[1.05pt]
\end{tabular*}

\vspace{2mm}

\caption{
\textbf{Generation quality (FID $\downarrow$) of pDiTs with and without register tokens on ImageNe 512.}
We compare generation quality across model sizes and training epochs.
Here, ``w/o regs.'' and ``w/ regs.'' denote training without and with register tokens, respectively.
\textcolor[HTML]{808080}{The shaded IC column reports performance with in-context class conditioning (Section~\ref{sec:incontext}).}
}
\label{tab:app_analysis_jit_registers_512}
\end{table*}

\subsection{Analysis on ImageNet}\label{app:analysis_imagenet}

\textbf{Outliers and registers in DiTs.} In the main text, we show that DiTs are free from the artifacts observed in ViTs. However, introducing register tokens leads to the emergence of high-norm tokens within the registers themselves. While the main analysis focuses on the small-scale pDiT-B/16 model at a single timestep ($t=0.5$), here we present corresponding results for additional timesteps, larger model variants, higher resolution, latent-space counterparts, and few-step models~\cite{lu2026one,geng2026mean}.

First, Figure~\ref{fig:app_analysis_jit_w_regs_more_t} shows that this effect consistently holds across different timesteps.

Second, Figure~\ref{fig:app_analysis_jit_wo_w_regs} shows that the observations made for pDiT-B/16 also hold for larger pixel-space variants, namely pDiT-L/16 and pDiT-H/16. Models without registers maintain relatively uniform patch-token norms across layers, without pronounced outliers. Once registers are introduced, however, large-norm tokens systematically emerge within the register tokens.

In Table~\ref{tab:app_analysis_jit_registers_512}, we show that registers also improve performance on ImageNet $512{\times}512$.

Additionally, we analyze where outliers emerge within each block. In Figure~\ref{fig:app_which_block}, we show the register-token norms after different sub-layers of several pDiT-B/16 blocks. We find that the outliers are mainly formed after the MLP layer rather than after the attention layer.

We analyze the latent-space models SiT~\cite{ma2024sit} and RAE~\cite{zheng2026diffusion} in terms of outlier behavior. Figures~\ref{fig:app_analysis_sit_wo_w_regs} and~\ref{fig:app_analysis_rae_wo_w_regs} show that the same phenomenon also holds for latent-space diffusion models.

Finally, we consider MeanFlows~\cite{geng2026improved} and pMF~\cite{lu2026one}, since they also use in-context conditioning in their implementations, and find that they also exhibit outliers within the in-context tokens. Figures~\ref{fig:app_imf_analysis} and~\ref{fig:app_pmf_analysis} show that the same phenomenon holds for few-step models in both latent and pixel spaces, across model sizes and image resolutions.

\textbf{Improving Feature Map Quality.}
Next, we find that register tokens consistently reduce the feature norms
of patch tokens. In Figure~\ref{fig:app_analysis_jit_norms}, we show this effect for larger pixel-space variants of pDiT. Interestingly, we do not observe the same behavior in SSL ViTs such as DINOv2, as shown in Figure~\ref{fig:app_analysis_dino_norms}.

We show that register tokens improve feature quality across larger pDiT variants. 
Figure~\ref{fig:app_analysis_jit_tv_ratio} reports the TV ratio for pDiT-L/16 and pDiT-H/16. 
In addition, Figure~\ref{fig:app_analysis_cka} analyzes the correlation decay slope~\cite{singh2025matters}, where lower values indicate stronger spatial organization. 
Both metrics show the same trend: register tokens improve internal representations at high noise levels, beginning at block $4$, where the registers are introduced.

\textbf{Effect of Register Injection Layer and Number of Register Tokens}. Our analysis shows that, unlike ViTs where register tokens are effective
from the first layer~\cite{darcet2023vision}, pDiT-B/16 benefits primarily from their delayed introduction. For instance, introducing registers across
all layers ($0$--$11$) results in performance similar to a model without registers.

Here, we provide a hypothesis for why this effect occurs.
In Figure~\ref{fig:app_analysis_probing}, we present linear probing results for register tokens introduced from layers $0$ and $4$.
Importantly, we observe a notable difference between these two
configurations. Specifically, models with registers introduced from the first layer produce substantially less informative register tokens. That is, we observe many tokens with moderate feature norms but very low probing accuracy. This suggests that these tokens serve neither as norm sinks nor as carriers of semantic information. Since we measure these results after the $5$th block, we hypothesize that this poor signal from register tokens can negatively affect later layers and consequently degrade model performance.

We hypothesize that this poor register signal arises because, in the early layers of pDiTs, the model has not yet formed meaningful semantic structure. As a result, register tokens cannot capture diverse semantic information and instead propagate uninformative features.

\begin{table}[t]
\centering
\small
\setlength{\tabcolsep}{8pt}
\renewcommand{\arraystretch}{1.15}
\begin{tabular}{lccc}
\toprule
& \textbf{RAE-space} & \textbf{VAE-space} & \textbf{Pixel-space} \\
& \textbf{pDiT backbone} & \textbf{pDiT backbone} & \textbf{pDiT backbone} \\
\midrule

\rowcolor{tabsection}
& \multicolumn{3}{c}{\textit{Base size}} \\
With registers
& \cellcolor{tabred} 10.12
& \cellcolor{tabamber} \textbf{7.17}
& \cellcolor{tabgreen} \textbf{5.30} \\
Without registers
& \cellcolor{tabred} \textbf{9.40}
& \cellcolor{tabamber} 7.20
& \cellcolor{tabgreen} 7.39 \\

\bottomrule
\end{tabular}
\vspace{3mm}

\caption{\textbf{Registers are more effective in pixel-space.}
    We compare the generation quality (FID) of models with and without register tokens across different training spaces (DINOv2, VAE, and pixel space) using the same pDiT backbone. 
    Register tokens provide the largest improvements in pixel space, moderate gains in VAE space, and degrade performance in DINOv2 space (RAE).
}
\label{tab:app_analysis_register_spaces}
\end{table}

\textbf{Pixel-space versus latent-space.} In the main text, we show that register tokens provide substantially larger gains for pixel-space models. In Table~\ref{tab:app_analysis_register_spaces}, we further present results for additional backbones in RAE and VAE spaces and observe the same trend: performance degrades for RAE-based models, while VAE-space models show similar performance with and without registers. This suggests that the effect is not backbone-specific.

Next, we provide an explanation for this behavior through an analysis of token feature norms and intermediate representations across different model types. In Figures~\ref{fig:app_analysis_norms_models} and ~\ref{fig:app_analysis_tvs_models}, we observe that pDiTs consistently exhibit the largest feature norms and the noisiest intermediate representations compared to latent-space counterparts. Specifically, Figure~\ref{fig:app_analysis_norms_models} shows that patch-token norms in pDiTs are substantially higher across all timesteps, while Figure~\ref{fig:app_analysis_tvs_models} demonstrates that their intermediate features have significantly larger TV values. 

These observations suggest that pixel-space diffusion produces substantially noisier intermediate features, which may explain why register tokens are especially beneficial in this setting.

\subsection{Analysis on Text-to-Image Models}\label{app:t2i}
In addition to ImageNet-based DiTs, we consider large-scale text-to-image approaches. We do not train these models with registers, but instead analyze pretrained open-source versions. Specifically, we analyze FLUX~\cite{flux2024}, SD3.5 Large~\cite{esser2024scaling}, Qwen-Image~\cite{wu2025qwen}, and Z-Image~\cite{cai2025z}, which propagate textual information through an auxiliary token sequence appended to the image tokens. Importantly, this sequence does not directly participate in the diffusion loss, raising the possibility that it may partially serve a register-like role. Moreover, we consider PixArt-$\alpha$~\cite{chen2024pixart}, which does not introduce an additional sequence into the image-token sequence and instead follows the classical DiT~\cite{albergo2025stochastic} regime with cross-attention.

We present the results in Figure~\ref{fig:app_analysis_t2i}. For SD3.5, Qwen-Image, and Z-Image, we observe that high-norm outliers emerge within the text-token sequence, while image-token norms remain comparatively uniform. This behavior closely resembles the role of register tokens in ImageNet-based DiTs, suggesting that auxiliary text tokens may implicitly act as repositories for high-norm representations.

Interestingly, for FLUX and PixArt-$\alpha$, we observe outliers not only in text tokens but also in several image tokens. However, these image-token outliers disappear in later layers, suggesting that patch tokens cannot sustain high norms due to the loss constraint. Moreover, in FLUX, their norms are smaller than those of the text-token outliers. In PixArt-$\alpha$, we find a single image-token outlier in an intermediate layer (block 16), which also disappears later.

These results suggest that the model needs space for outliers, but patch tokens are not convenient slots for this role. This further supports the importance of registers as dedicated outlier tokens.

\section{Additional Experiments} \label{app:additional_exps}
\subsection{Register Guidance}
\label{app:rg_results}

\begin{table*}[t!]
\centering
\small
\setlength{\tabcolsep}{3pt}
\renewcommand{\arraystretch}{1.12}

\begin{minipage}[t]{0.49\textwidth}
\centering
\begin{tabular*}{\linewidth}{@{\extracolsep{\fill}}lccccc@{}}
\toprule[1.2pt]
$p=0.0$
& epoch $=0$ & $10$ & $20$ & $30$ & $50$ \\
\midrule
CFG
& $3.60$ & $3.46$ & $3.41$ & $3.44$ & ${3.38}$ \\
RG
& $13.13$ & $13.36$ & $13.88$ & $14.03$ & $14.19$ \\
RG+CFG
& $8.02$ & $8.17$ & $8.20$ & $8.43$ & $8.41$ \\
\bottomrule[1.2pt]
\end{tabular*}
\end{minipage}
\hfill
\begin{minipage}[t]{0.49\textwidth}
\centering
\begin{tabular*}{\linewidth}{@{\extracolsep{\fill}}lccccc@{}}
\toprule[1.2pt]
$p=0.1$
& epoch $=0$ & $25$ & $30$ & $50$ & $85$ \\
\midrule
CFG
& $3.60$ & $3.55$ & $3.51$ & $3.44$ & $3.41$ \\
RG
& $13.13$ & $5.33$ & $4.89$ & $3.94$ & $3.48$ \\
RG+CFG
& $8.02$ & $3.51$ & $3.29$ & $3.10$ & $2.90$ \\
\bottomrule[1.2pt]
\end{tabular*}
\end{minipage}

\vspace{4mm}

\begin{minipage}[t]{0.49\textwidth}
\centering
\begin{tabular*}{\linewidth}{@{\extracolsep{\fill}}lccccc@{}}
\toprule[1.2pt]
$p=0.3$
& epoch $=0$ & $10$ & $20$ & $30$ & $50$ \\
\midrule
CFG
& $3.60$ & $3.59$ & $3.61$ & $3.59$ & $3.52$ \\
RG
& $13.13$ & $6.96$ & $4.74$ & $3.65$ & $3.25$ \\
RG+CFG
& $8.02$ & $4.41$ & $3.38$ & $3.15$ & $2.96$ \\
\bottomrule[1.2pt]
\end{tabular*}
\end{minipage}
\hfill
\begin{minipage}[t]{0.49\textwidth}
\centering
\begin{tabular*}{\linewidth}{@{\extracolsep{\fill}}lccccc@{}}
\toprule[1.2pt]
$p=0.7$
& epoch $=0$ & $10$ & $20$ & $30$ & $50$ \\
\midrule
CFG
& $3.60$ & $3.79$ & $3.81$ & $3.79$ & $3.68$ \\
RG
& $13.13$ & $6.72$ & $4.24$ & $3.50$ & $3.95$ \\
RG+CFG
& $8.02$ & $4.45$ & $3.51$ & $3.29$ & $3.21$ \\
\bottomrule[1.2pt]
\end{tabular*}
\end{minipage}

\vspace{2mm}

\caption{
\textbf{Fine-tuning results for JiT-B/16 with different register-drop probabilities $p$.}
We fine-tune a pretrained JiT-B/16 checkpoint and report FID across epochs for CFG, RG, and RG+CFG.
$p=0$ corresponds to fine-tuning the original model with the basic setup, which does not improve RG.
A low value, $p=0.1$, requires longer convergence, while a high value, $p=0.7$, degrades the original performance.
Therefore, we use the intermediate value $p=0.3$.
}
\label{tab:app_exps_rg_ablations}
\end{table*}

\begin{table*}[t!]
\centering
\small
\setlength{\tabcolsep}{5pt}
\renewcommand{\arraystretch}{1.12}

\begin{tabular*}{\textwidth}{@{\extracolsep{\fill}}lcccccc@{}}
\toprule[1.2pt]
& \multicolumn{3}{c}{$p=0.0$} & \multicolumn{3}{c}{$p=0.3$} \\
\cmidrule(lr){2-4}
\cmidrule(lr){5-7}
Model
& CFG & RG & RG+CFG
& CFG & RG & RG+CFG \\
\midrule

JiT-B/32
& $3.77$ & $12.61$ & $7.75$
& $3.90$ & $3.62$ & $\mathbf{3.34}$ \\

JiT-L/16
& $2.30$ & $7.82$ & $4.50$
& $2.39$ & $\mathbf{2.00}$ & $2.13$ \\

JiT-L/32
& $2.55$ & $12.06$ & $8.05$
& $2.57$ & $2.24$ & $\mathbf{2.23}$ \\

JiT-H/16
& $1.90$ & $7.67$ & $3.68$
& $1.92$ & $1.85$ & $\mathbf{1.80}$ \\

JiT-H/32
& $1.95$ & $13.37$ & $6.76$
& $1.94$ & $2.01$ & $\mathbf{1.87}$ \\

\bottomrule[1.2pt]
\end{tabular*}

\vspace{1mm}

\caption{
\textbf{Best fine-tuning results for different JiT models.}
We report the best FID across fine-tuning epochs for each setting.
Register-drop fine-tuning with $p=0.3$ enables effective RG and RG+CFG, outperforming the original fine-tuning setup with $p=0.0$ using CFG.
As expected, $p=0.0$ does not enable effective RG.
}
\label{tab:app_exps_rg_best_finetuning}
\end{table*}

In the main text, we find that Register Guidance produces artifacts when using two separate models, despite the fact that they generate similar images. However, when using a single model that can generate both with and without registers, this mismatch disappears and no artifacts are introduced. Here, we provide additional visual results for the two approaches in Figures~\ref{fig:app_uncond_guidance_separate_full} and~\ref{fig:app_uncond_guidance_full}.

In our experiments, we find that the best configuration combines CFG and RG. Here, we provide visual results illustrating this behavior in Figure~\ref{fig:app_cfg_rg_visual}.

To achieve the best results, we run our approach in the fine-tuning regime. Specifically, we take pretrained JiT checkpoints and fine-tune them for several epochs. We use a larger register-drop probability $p$ than in the training-from-scratch setup. In Table~\ref{tab:app_exps_rg_ablations}, we ablate different values of $p$ for JiT-B/16 and report FID at different fine-tuning epochs. As expected $p=0$ does not enable effective RG, while $p=0.1$ requires longer fine-tuning to converge. A large value, $p=0.7$, improves RG but degrades the CFG baseline. The intermediate value $p=0.3$ provides the best trade-off, yielding strong RG performance while preserving the original model quality. In Table~\ref{tab:app_exps_rg_best_finetuning}, we report results for other models using the selected value $p=0.3$ and their optimal RG scales.

\subsection{Decoupled Processing of Register and Patch Tokens}
\label{app:dual}

\begin{table}[t!]
\centering

\begin{minipage}[t]{0.315\linewidth}
\hrule height 0.8pt
\vspace{0.25em}

{\snippettitle{snip:rmsnorm}{RMSNorm Dual}}

\vspace{0.25em}
\hrule height 0.5pt
\vspace{0.35em}

\lstset{
  language=trainalgo,
  basicstyle=\ttfamily\fontsize{7.5}{8.5}\selectfont,
  showstringspaces=false,
  frame=none,
  xleftmargin=0pt,
  xrightmargin=0pt,
  aboveskip=0pt,
  belowskip=0pt,
  keepspaces=true,
  columns=fullflexible,
  breaklines=true
}
\begin{lstlisting}
# registers + patches
# x: [B, n_reg + n_patch, h] 

# dual RMSNorm
(x_reg, x_patch) = split(x)

# separate normalization
x_patch = RMSNorm(x_patch, w1, eps1)
x_reg = RMSNorm(x_reg, w2, eps2)

# merge streams
x = concat(x_reg, x_patch)
\end{lstlisting}

\vspace{0.25em}
\hrule height 0.5pt
\end{minipage}
\hfill
\begin{minipage}[t]{0.315\linewidth}
\hrule height 0.8pt
\vspace{0.25em}

{\snippettitle{snip:swiglu}{SwiGLU MLP Dual}}

\vspace{0.25em}
\hrule height 0.5pt
\vspace{0.35em}

\lstset{
  language=trainalgo,
  basicstyle=\ttfamily\fontsize{7.5}{8.5}\selectfont,
  showstringspaces=false,
  frame=none,
  xleftmargin=0pt,
  xrightmargin=0pt,
  aboveskip=0pt,
  belowskip=0pt,
  keepspaces=true,
  columns=fullflexible,
  breaklines=true
}
\begin{lstlisting}
# shared hidden latent
x1, x2 = chunk(Linear(x))
hidden = silu(x1) * x2

# split output projection
(h_reg, h_patch) = split(hidden)
y_reg   = Linear(h_reg)
y_patch = Linear(h_patch)

# merge streams
y = concat(y_reg, y_patch)
\end{lstlisting}

\vspace{0.25em}
\hrule height 0.5pt
\end{minipage}
\hfill
\begin{minipage}[t]{0.315\linewidth}
\hrule height 0.8pt
\vspace{0.25em}

{\snippettitle{snip:adaln}{adaLN Dual}}

\vspace{0.25em}
\hrule height 0.5pt
\vspace{0.35em}

\lstset{
  language=trainalgo,
  basicstyle=\ttfamily\fontsize{7.5}{8.5}\selectfont,
  showstringspaces=false,
  frame=none,
  xleftmargin=0pt,
  xrightmargin=0pt,
  aboveskip=0pt,
  belowskip=0pt,
  keepspaces=true,
  columns=fullflexible,
  breaklines=true
}
\begin{lstlisting}
# condition
# c: [B, h] 

# shared modulation
m = Linear(silu(c))

# dual branch params
m_r = m + LoRA(c)
(shift, scale, gate) = split(m)
(shift_reg, scale_reg, gate_reg) = split(m_r)
\end{lstlisting}

\vspace{0.25em}
\hrule height 0.5pt
\end{minipage}

\end{table}

\textbf{Dual-stream architecture.} Our analysis reveals that register and patch tokens play distinct roles, yet existing pixel-space architectures~\cite{li2025back, lu2026one} process them identically using fully shared parameters. Given their differing behaviors, such parameter sharing may be suboptimal.

In contrast, large-scale models with appended auxiliary sequences (e.g., text tokens) often employ dual-stream architectures~\cite{esser2024scaling, flux2024}, where separate parameters are used for different token types while interactions are maintained through attention. Motivated by these differences, we investigate dual-stream designs that enable specialized processing of register tokens in pixel-space DiTs.

We consider the JiT architecture~\cite{li2025back}, which uses in-context conditioning. JiT blocks consist of \verb+RMSNorm+, \verb+adaLN+, \verb+Attention+, and \verb+MLP+ layers. Thus, we selectively introduce token-specific specialization in these components.

In Table~\ref{tab:exps_full_dual}, we evaluate naive dual-stream designs that separately duplicate transformer components for register and patch tokens. Duplicating all components gives the best performance but significantly increases parameters. Among individual components, \verb+adaLN+ and \verb+MLP+ contribute the most, while \verb+Attention+ and \verb+RMSNorm+ provide limited gains.

\textbf{Compact dual-stream design.} As we find, naive duplication leads to a significant increase in parameters. We therefore explore parameter-efficient strategies for selective duplication.

For \verb+RMSNorm+, which has few parameters, we use separate parameters for register and patch tokens with negligible cost (Snippet~\ref{snip:rmsnorm}).

For parameter-intensive modules (\verb+adaLN+, \verb+Attention+, and \verb+MLP+), we avoid full duplication. 
In \verb+MLP+, we compute a shared \verb+SwiGLU+ (\verb+Linear+ projection followed by \verb+SiLU+ gating) and apply separate output projections for register and patch tokens (Snippet~\ref{snip:swiglu}).  

For \verb+adaLN+ and \verb+Attention+, we use parameter-efficient LoRA adaptations~\cite{hu2022lora}, following~\cite{marouani2026revisiting}. Shared parameters are computed for all tokens, while lightweight LoRA branches are applied only to register tokens. In \verb+adaLN+, normalization parameters are first shared across all tokens, then refined for register tokens via a LoRA branch (Snippet~\ref{snip:adaln}). In \verb+Attention+, QKV projections are shared, with LoRA applied only to register-token representations, while the output projection remains shared.

Our final architecture combines single-stream and dual-stream transformer layers. 
Motivated by the ablation results in Table~\ref{tab:analysis_register_start}, which show that register tokens are ineffective in early layers, we first process the image sequence using standard single-stream layers. Register tokens are then introduced only in later stages, where the model transitions to dual-stream processing. The overall design increases the parameter count by only ${\sim}14\%$, compared to ${\sim}77\%$ for naive duplication.

\begin{table*}[t]
\centering
\small
\renewcommand{\arraystretch}{1.12}
\newcommand{\cmark}{\checkmark}

\begin{minipage}[t]{0.48\textwidth}
\centering
\setlength{\tabcolsep}{4pt}

\resizebox{\linewidth}{!}{%
\begin{tabular}{@{}cccccc@{}}
\toprule
adaLN & MLP & Attn & RMSNorm & Params (M) & FID \\
\midrule

\rowcolor[HTML]{eeeeee}
& & & & $131$M & $3.71$ \\

\cmark & \cmark & \cmark & \cmark & $230$M & $\mathbf{3.14}$ \\

\cmark &  &  & & $173$M & ${3.44}$ \\

 &  \cmark &  &  & $168$M & $3.44$ \\

 &  &  \cmark & & $150$M & $3.61$ \\

 &  &  &  \cmark & $131$M & $3.70$ \\

\bottomrule
\end{tabular}
}
\vspace{1mm}
\caption{
\textbf{Ablation of dual-stream designs.} We separately duplicate transformer components for register and patch tokens. Duplicating all components gives the best performance but significantly increases parameters. Among individual components, \texttt{adaLN} and \texttt{MLP} contribute the most, while \texttt{Attention} and \texttt{RMSNorm} provide limited gains. \textcolor[HTML]{808080}{gray} denotes the single-stream JiT-B/16 baseline.
}

\label{tab:exps_full_dual}
\end{minipage}
\hfill
\begin{minipage}[t]{0.48\textwidth}
\vspace{-15.6mm}
\centering
\setlength{\tabcolsep}{4pt}
\resizebox{\linewidth}{!}{%
\begin{tabular}{@{}cccccc@{}}
\toprule
adaLN & MLP & Attn & RMSNorm & Params (M) & FID \\
\midrule

\rowcolor[HTML]{eeeeee}
& & & & $131$M & $3.71$ \\

\cmark & \cmark & \cmark & \cmark & $161$M & $3.48$ \\

\rowcolor[HTML]{EEF9F1}
\cmark & \cmark &  & \cmark & $149$M & $\mathbf{3.41}$ \\

\cmark &  &  & \cmark & $136$M & $3.81$ \\

 & \cmark &  & \cmark & $143$M & $3.56$ \\

\cmark & \cmark &  &  & $149$M & $3.53$ \\

\bottomrule
\end{tabular}
}
\vspace{1mm}
\caption{
\textbf{Ablation of compact dual-stream designs.} We evaluate parameter-efficient dual-stream configurations by selectively dualizing transformer components for register and patch tokens. The best compact design (\textcolor[HTML]{3A8F5B}{green}) dualizes \texttt{adaLN}, \texttt{MLP}, and \texttt{RMSNorm}, while keeping \texttt{Attention} shared, achieving the best trade-off between parameter count and generation quality. 
}
\label{tab:exps_compact_dual}
\end{minipage}

\end{table*}

\begin{table}[t!]
\centering

\begin{minipage}{0.46\textwidth}
\centering

\begin{adjustbox}{width=1.0\textwidth}
\begin{tabular}{@{} l c c c c @{}}
\toprule
Model & Params (M) & GFLOPs & Epoch & FID \\
\midrule

\multirow{2}{*}{JiT-B/16} & 
 \multirow{2}{*}{$131$} & \multirow{2}{*}{$23.8$} & $200$ & $4.71$ \\
  & & & $600$ & $3.71$\vspace{2mm}\\
 
\multirow{2}{*}{JiT-L/16} 
 & \multirow{2}{*}{$459$} & \multirow{2}{*}{$84.2$} & $200$ & $2.95$ \\
& & & $600$ & $2.47$ \\
\midrule

\multirow{2}{*}{Ours-B/16} 
& \multirow{2}{*}{$149$} & \multirow{2}{*}{$23.8$} & $200$ & $\mathbf{4.25}$ \\
 &  &  & $600$ & $\mathbf{3.41}$\vspace{2mm}\\

\multirow{2}{*}{Ours-L/16} 
& \multirow{2}{*}{$518$} & \multirow{2}{*}{$84.2$} & $200$ & $\mathbf{2.76}$ \\
&  &  & $600$ & $\mathbf{2.32}$ \\

\bottomrule
\end{tabular}
\end{adjustbox}

\vspace{1mm}

\caption{
    Comparison of our compact dual-stream architecture with the JiT baseline on ImageNet $256{\times}256$.
}
\label{tab:exps_imagenet256}

\end{minipage}
\hspace{0.04\textwidth}
\begin{minipage}{0.46\textwidth}
\centering
\begin{adjustbox}{width=1.0\textwidth}
\begin{tabular}{@{} l c c c c @{}}
\toprule
Model & Params (M) & GFLOPs & Epoch & FID \\
\midrule

\multirow{2}{*}{JiT-B/32}
 & \multirow{2}{*}{$133$} & \multirow{2}{*}{$24.3$} & $200$ & $5.84$ \\
&  &  & $600$ & $4.12$\vspace{2mm}\\
 
\multirow{2}{*}{JiT-L/32} 
 & \multirow{2}{*}{$461$} & \multirow{2}{*}{$84.8$} & $200$ & $3.28$ \\
 &  & & $600$ & $2.69$ \\
\midrule

\multirow{2}{*}{Ours-B/32} 
 & \multirow{2}{*}{$151$} & \multirow{2}{*}{$24.3$} & $200$ & $\mathbf{4.98}$ \\
 & &  & $600$ & $\mathbf{3.92}$\vspace{2mm}\\

\multirow{2}{*}{Ours-L/32} 
 & \multirow{2}{*}{$520$} & \multirow{2}{*}{$84.8$} & $200$ & $\mathbf{3.08}$ \\
 & & & $600$ & $\mathbf{2.57}$ \\

\bottomrule
\end{tabular}
\end{adjustbox}
\vspace{1mm}
\caption{
    Comparison of our compact dual-stream architecture with the JiT baseline on ImageNet $512{\times}512$.
}
\label{tab:exps_imagenet512}
\end{minipage}
\end{table}

\textbf{Results}. We consider models of two sizes, B and L, and train them on ImageNet at resolutions of 256 and 512. The proposed architecture introduces an additional parameter overhead of approximately $14\%$. We train the models using the same training and inference configuration as in JiT~\cite{li2025back}. For both configurations, we use LoRA with rank $128$ in AdaLN.

We first study which components should be decoupled in the compact dual-stream design.
We use JiT-B/16 adapted to a dual-stream architecture and train it on ImageNet $256\times256$ for $600$ epochs, following the training and sampling setups of~\cite{li2025back}. 
We enable in-context tokens at layer $4$, resulting in single-stream layers $0$--$3$ and dual-stream layers $4$--$11$.

Table~\ref{tab:exps_compact_dual} reports the results for several parameter-efficient dual-stream designs, compared against the single-stream JiT-B/16 baseline (\textcolor[HTML]{808080}{gray-shaded row}).

Applying the dual architecture to all layers improves FID from $3.71$ to $3.48$. Sharing \verb+Attention+ further improves performance while reducing parameters, suggesting that attention decoupling is unnecessary. In contrast, sharing \verb+MLP+ significantly degrades performance, while removing dual \verb+adaLN+ or \verb+RMSNorm+ also hurts results. Based on these findings, our final compact design dualizes \verb+MLP+, \verb+adaLN+, and \verb+RMSNorm+, while keeping \verb+Attention+ shared (\textcolor[HTML]{4F9D69}{green-shaded row}).

Next, we conduct a system-level comparison between our compact dual architecture and the JiT baseline on ImageNet $256\times256$ and $512\times512$ across different model sizes. 
For all variants, we follow the training setup and baseline configurations of~\cite{li2025back}.

We report the results in Tables~\ref{tab:exps_imagenet256} and~\ref{tab:exps_imagenet512}. The parameter-efficient dual-stream architecture consistently improves generation quality across model scales and image resolutions, while introducing only a small increase in parameters and no additional runtime overhead in terms of GFLOPs.

We believe our compact strategy can provide greater benefits at larger scales, where dual-stream architectures~\cite{esser2024scaling} achieve significant quality improvements. In these settings, the appended sequence, e.g., text, becomes substantially larger than the small set of register tokens.

\subsection{Other results}
\label{app:other}

\begin{table}[t!]
\centering
\label{tab:jit_b_results}
\begin{tabular}{lcc}
\toprule
\textbf{Method} & 200 epochs & 600 epochs \\
\midrule
pDiT-B/16 + PixelREPA     & $\mathbf{4.15}$   & $\mathbf{3.38}$ \\
pDiT-B/16 (w/o registers) + PixelREPA   & $4.78$    & $3.56$ \\
\midrule
JiT-B/16 + PixelREPA + compact dual     & $\mathbf{3.68}$   & $\mathbf{3.23}$ \\
JiT-B/16 + PixelREPA                    & $4.00$   & $3.40$ \\
\bottomrule
\end{tabular}
\vspace{2mm}
\caption{
\textbf{REPA remains complementary to register-like tokens and further benefits from the proposed compact dual-stream design.} We evaluate pDiT models with and without register tokens using PixelREPA~\cite{shin2026representation} on ImageNet $256{\times}256$. Register-like tokens consistently improve performance, and combining them with the proposed compact dual-stream architecture further improves the results.
}
\label{tab:app_exps_additional}
\end{table}

\begin{table}[t]
\centering
\setlength{\tabcolsep}{6pt}
\renewcommand{\arraystretch}{1.1}

\begin{tabular}{l c c c c c c}
\toprule
 & \multicolumn{3}{c}{Registers configuration} & \multicolumn{3}{c}{FID at Epoch} \\
\cmidrule(lr){2-4} \cmidrule(lr){5-7}
 & Size & Start & End & 40 & 80 & 120 \\
\midrule

\multirow{6}{*}{w/ reg.}
& $32$ & $4$ & $11$ & $\mathbf{37.7}$ & $\mathbf{9.59}$ & $\mathbf{6.45}$ \\
& $32$ & $4$ & $9$  & $36.9$ & $9.95$ & $6.65$ \\
& $32$ & $0$ & $11$ & $59.7$ & $19.3$ & $11.9$ \\
& $32$ & $0$ & $4$  & $62.4$ & $19.6$ & $12.3$ \\
& $16$ & $4$ & $11$ & $40.4$ & $10.2$ & $6.80$ \\
& $16$ & $0$ & $11$ & $62.4$ & $18.6$ & $11.3$ \\
& $4$  & $4$ & $11$ & $46.3$ & $12.8$ & $8.37$ \\
& $4$  & $0$ & $11$ & $54.8$ & $16.2$ & $10.4$ \\

\midrule

w/o reg.
& $-$ & $-$ & $-$ & $60.6$ & $18.4$ & $11.1$ \\

\bottomrule
\end{tabular}

\vspace{2mm}
\caption{
\textbf{Registers are effective in deeper layers.}
pDiT-B/16 benefits from registers only after the first $4$ layers; early-layer registers match the no-register baseline. More registers improve quality.
}
\label{tab:app_analysis_register_start}

\end{table}

In Table~\ref{tab:app_exps_additional}, we evaluate whether register-like tokens remain beneficial when combined with representation alignment~\cite{yu2024representation}. 
We adopt the recent REPA adaptation~\cite{shin2026representation}, which was specifically proposed for JiT-based architectures. We observe that register tokens consistently improve the performance of REPA-enhanced models. Moreover, our compact dual-stream design remains effective with REPA.

In addition, we ablate register-token configurations in Table~\ref{tab:app_analysis_register_start}, varying both the number of registers and the blocks where they are enabled. The results show that delayed introduction is crucial: enabling registers only in deeper layers ($4$--$11$) consistently outperforms both the no-register baseline and using registers from the first layer ($0$--$11$). Increasing the number of registers also improves performance, with $32$ registers performing best. Comparing the $4$--$11$ and $4$--$9$ settings further suggests that the latest layers contribute relatively little to register effectiveness.

In Figures~\ref{fig:app_rg_cfg_imgs_1} and~\ref{fig:app_rg_cfg_imgs_2}, we present qualitative generation results of our approach, where CFG and RG are used together for the JiT-H/16 model, achieving an FID of $1.80$.

\section{Limitations}
Our study focuses mainly on ImageNet DiTs and a set of pretrained text-to-image models. While the observed behavior is consistent across architectures, model sizes, resolutions, and few-step models, broader evaluation across more training recipes and datasets would further strengthen the conclusions. Although we show that registers improve intermediate representations and are closely connected to generation quality, we do not provide a comprehensive explanation of why they help. 

Register Guidance also introduces additional hyperparameters, such as the register-drop probability $p$ and guidance scale $w_{\mathrm{rg}}$, which may need to be tuned for new tasks, datasets, or architectures. Moreover, the best-performing RG+CFG setup requires additional model evaluations, increasing the number of function evaluations and limiting sampling speed compared to the original model.

\clearpage

\begin{figure*}[t!]
    \centering
    \includegraphics[width=\linewidth]{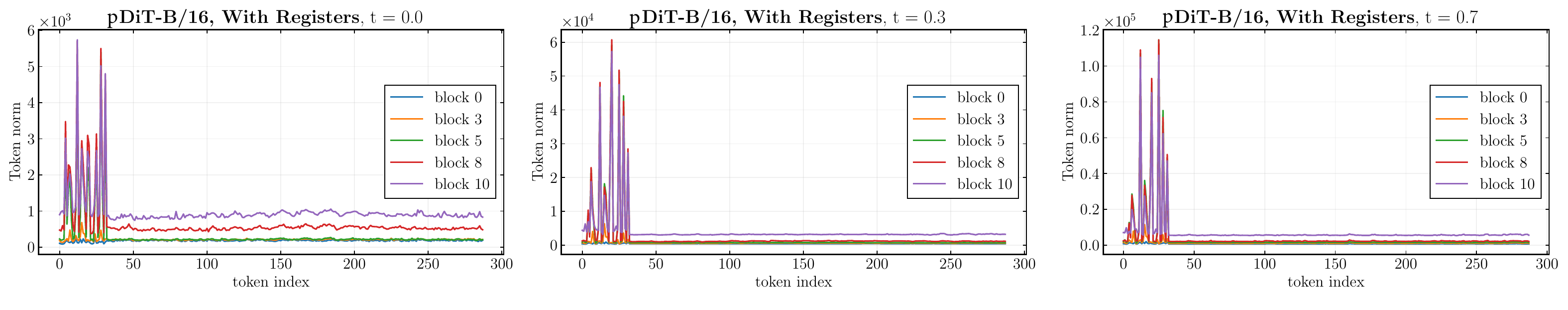}
    \caption{\textbf{High-norm outliers consistently emerge within register tokens across timesteps.} We visualize token-wise feature norms of pDiT-B/16 with registers for $t=0.0$, $0.3$, and $0.7$, and observe the same behavior in all cases.}
    \label{fig:app_analysis_jit_w_regs_more_t}
\end{figure*}

\begin{figure*}[t!]
    \centering
    \includegraphics[width=1.0\linewidth]{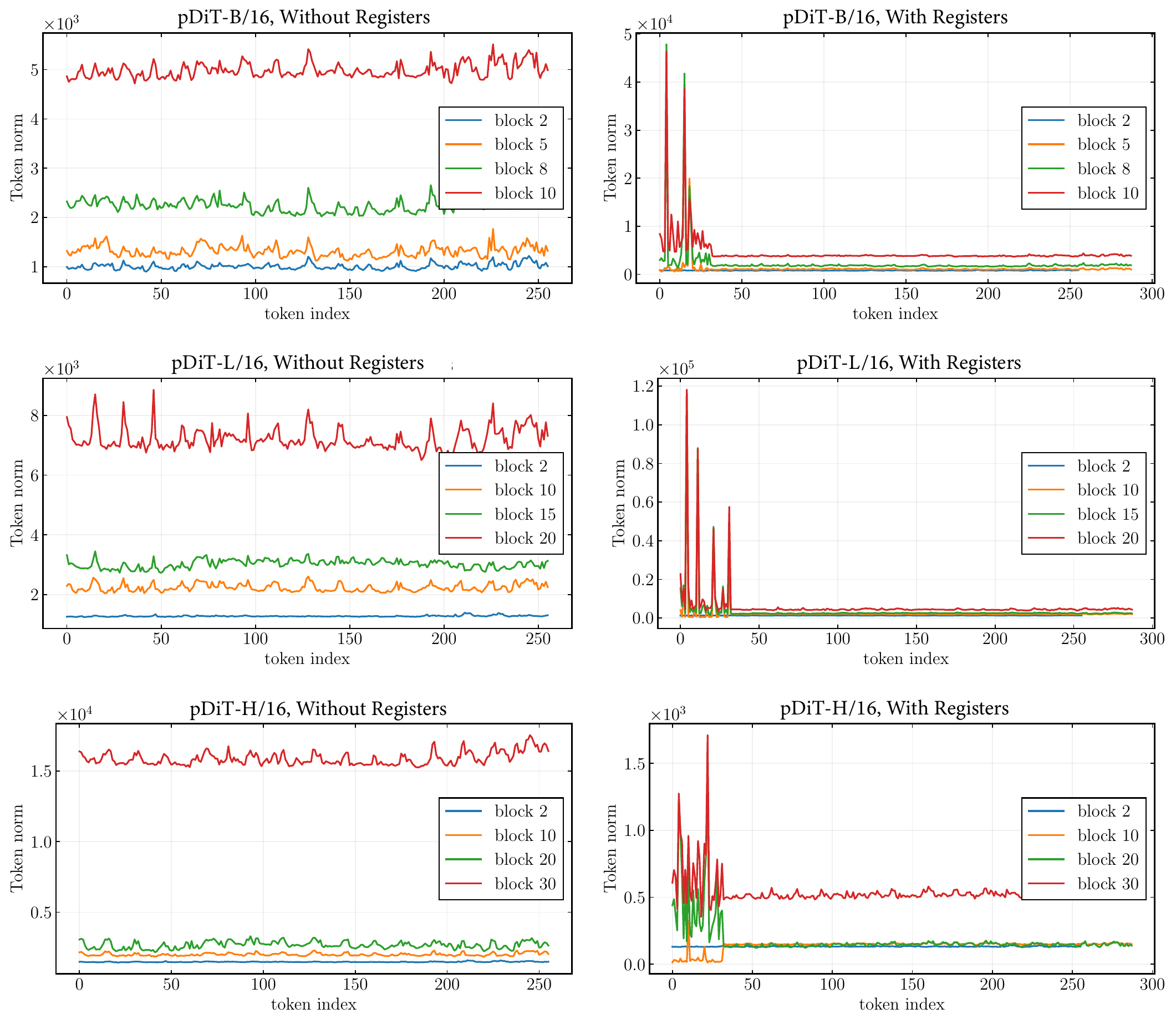}
    \caption{\textbf{Token-wise feature norms for pDiTs of varying scales on ImageNet $256\times256$, with and without registers.} Without registers, patch-token norms remain uniform across scales. Introducing registers leads to the emergence of high-norm outliers within the register tokens.}
    \label{fig:app_analysis_jit_wo_w_regs}
\end{figure*}

\begin{figure*}[t!]
    \centering
    \includegraphics[width=1.0\linewidth]{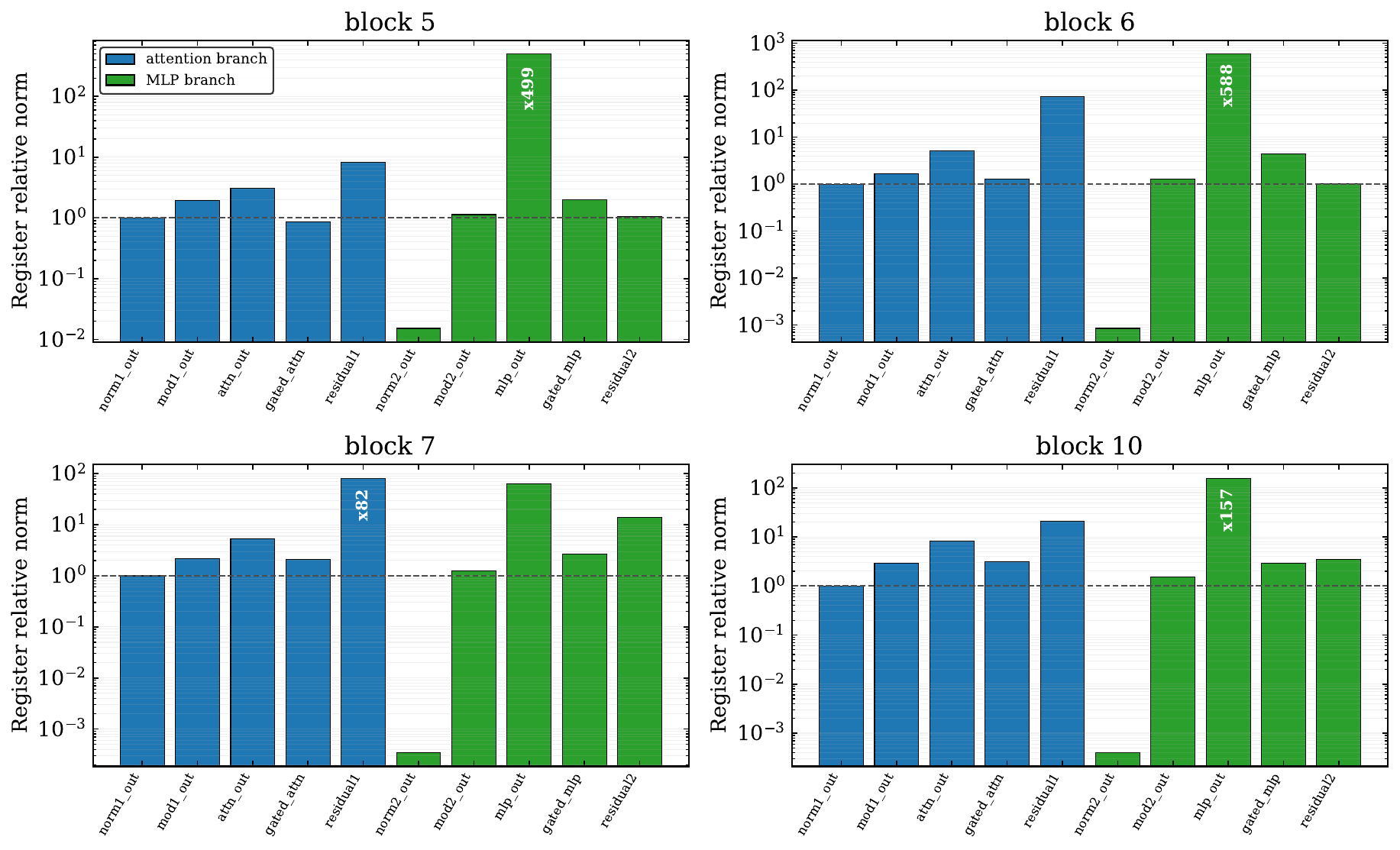}
    \caption{
\textbf{High-norm register tokens are amplified mainly inside the MLP branch.}
For blocks $5,6,7,10$ of pDiT-B/16 on ImageNet $256{\times}256$, we show the relative change in register-token feature norms across intra-block sub-layers.
Attention and MLP sub-layers are shown separately.
The largest norm amplification occurs at the MLP output, reaching up to $\times 499$ in block $5$, whereas the attention branch produces much smaller changes.
This indicates that register-token outliers are primarily formed by the MLP rather than by attention.
}
    \label{fig:app_which_block}
\end{figure*}

\begin{figure*}[t!]
    \centering
    \includegraphics[width=1.0\linewidth]{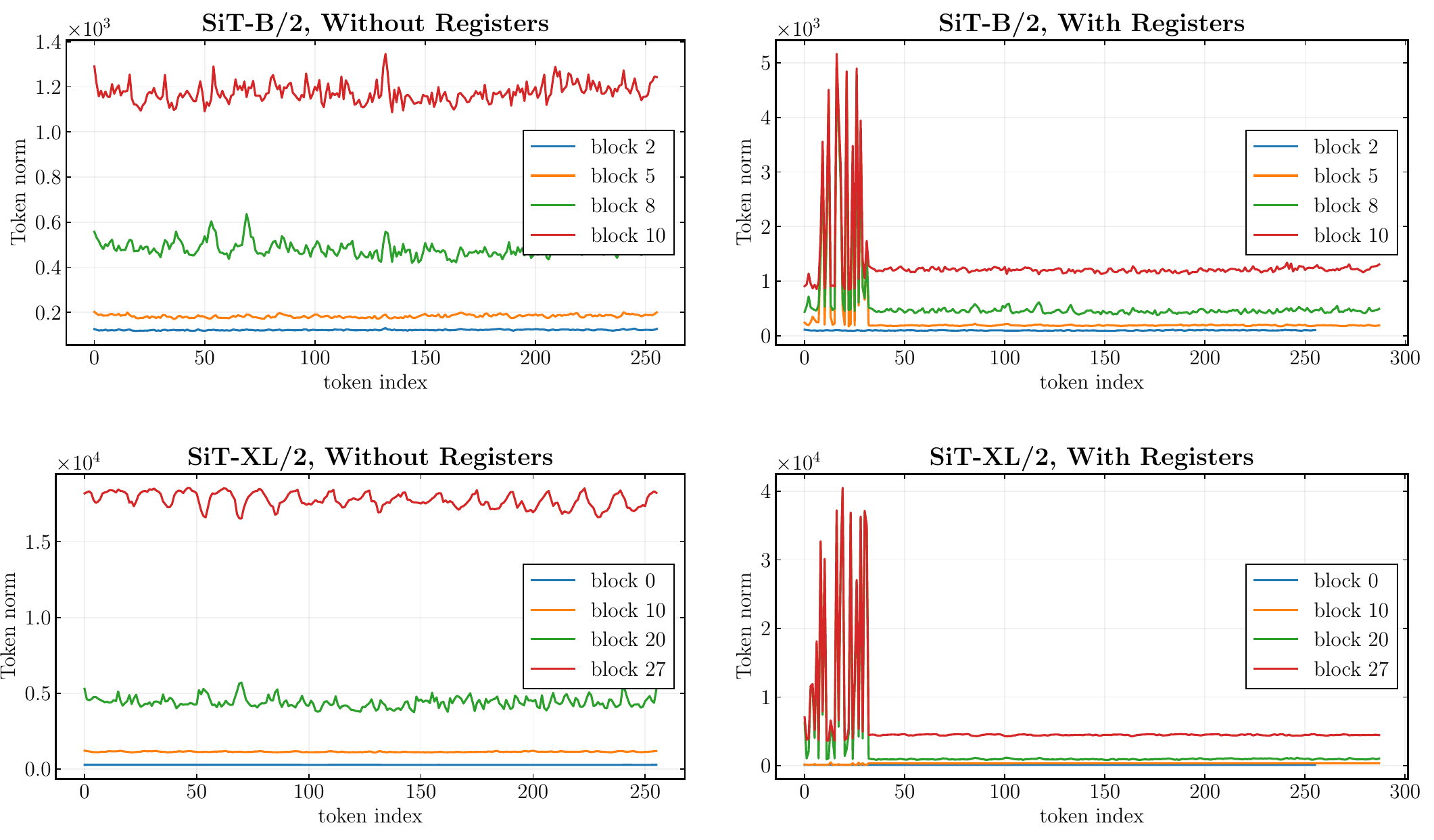}
    \caption{\textbf{Token-wise feature norms for VAE-space SiTs of varying scales on ImageNet $256\times256$, with and without registers.} SiTs without registers exhibit uniform patch-token norms across scales, while adding registers produces high-norm register tokens.}
    \label{fig:app_analysis_sit_wo_w_regs}
\end{figure*}

\begin{figure*}[t!]
    \centering
    \includegraphics[width=0.95\linewidth]{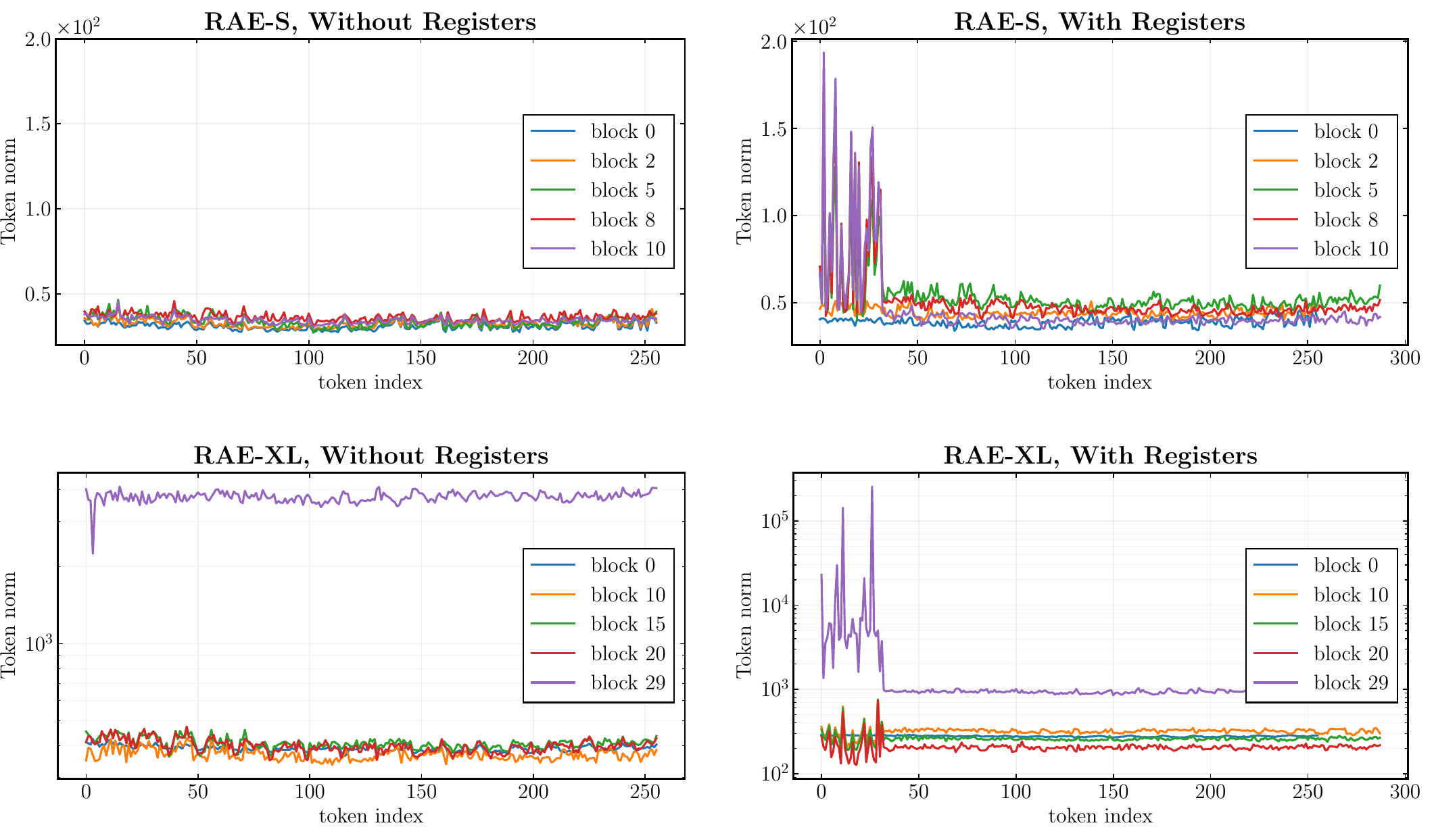}
    \caption{\textbf{Token-wise feature norms for DINOv2-space RAEs of varying scales on ImageNet $256\times256$, with and without registers.} RAEs without registers exhibit uniform patch-token norms across scales, while adding registers produces high-norm register tokens.}
    \label{fig:app_analysis_rae_wo_w_regs}
\end{figure*}

\begin{figure*}[t!]
    \centering
    \includegraphics[width=0.95\linewidth]{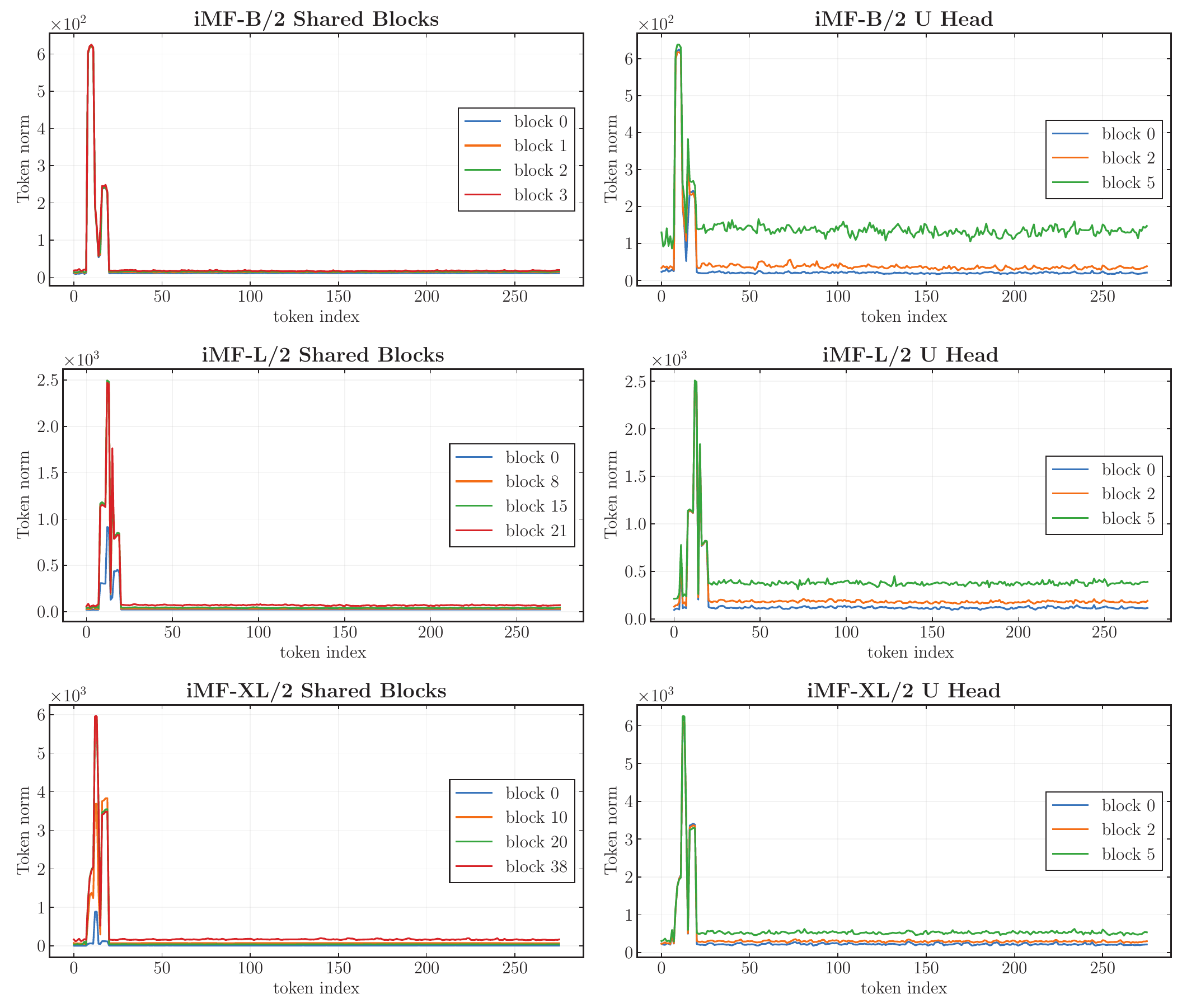}
    \caption{
    \textbf{Token-wise feature norms for improved MeanFlows~\cite{geng2026improved} of varying scales on ImageNet $256{\times}256$ with in-context conditioning.}
    We consider both shared blocks and blocks that predict the average velocity, and find high-norm tokens within the in-context tokens.
    }
    \label{fig:app_imf_analysis}
\end{figure*}

\begin{figure*}[t!]
    \centering
    \includegraphics[width=0.9\linewidth]{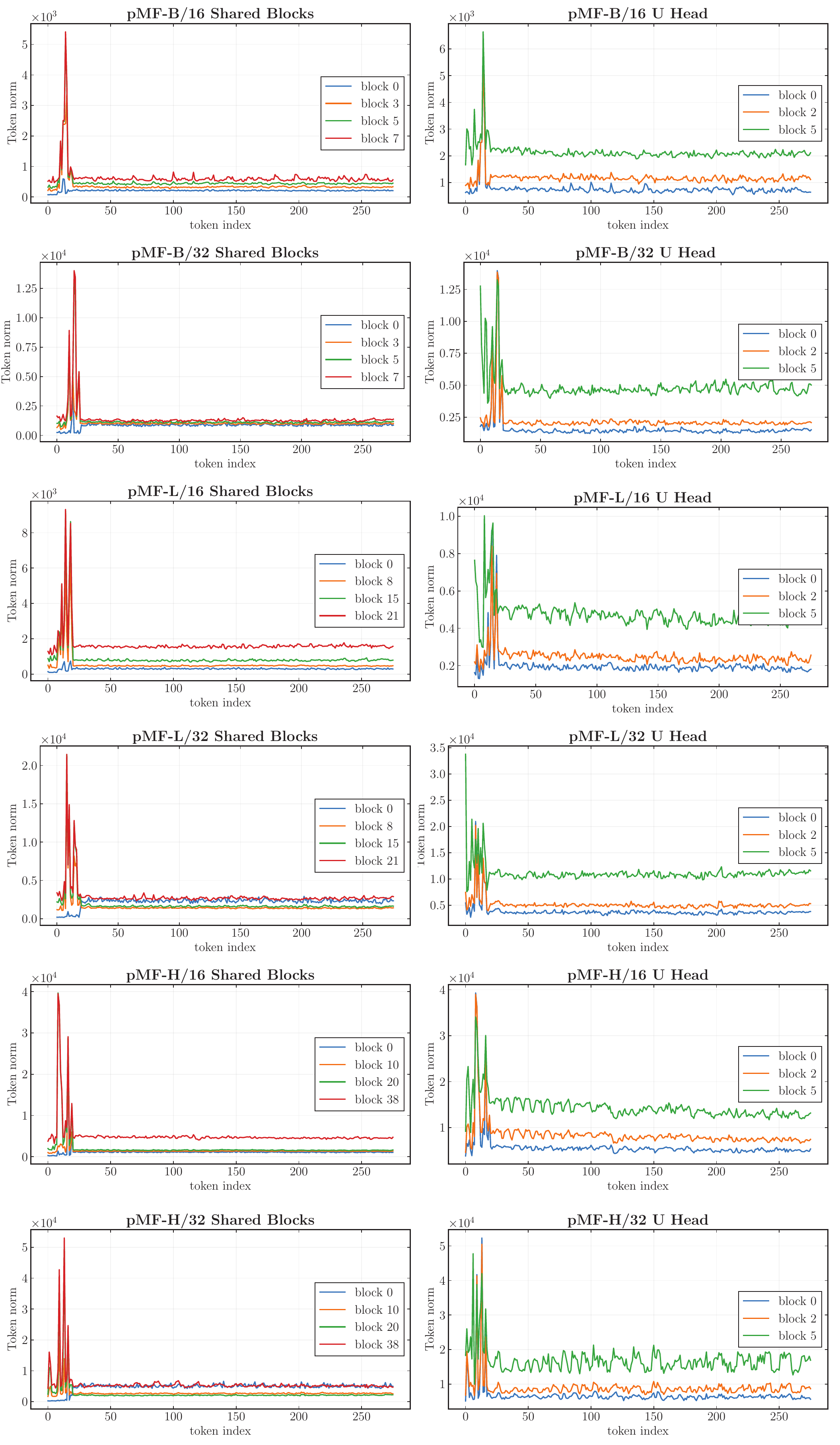}
   \caption{
    \textbf{Token-wise feature norms for pixel MeanFlows~\cite{lu2026one} of varying scales on ImageNet $256{\times}256$ and $512{\times}512$ with in-context conditioning.}
    We consider both shared blocks and blocks that predict the average velocity, and find high-norm tokens within the in-context tokens.
    }
    \label{fig:app_pmf_analysis}
\end{figure*}

\begin{figure*}[t!]
    \centering
    \includegraphics[width=1.0\linewidth]{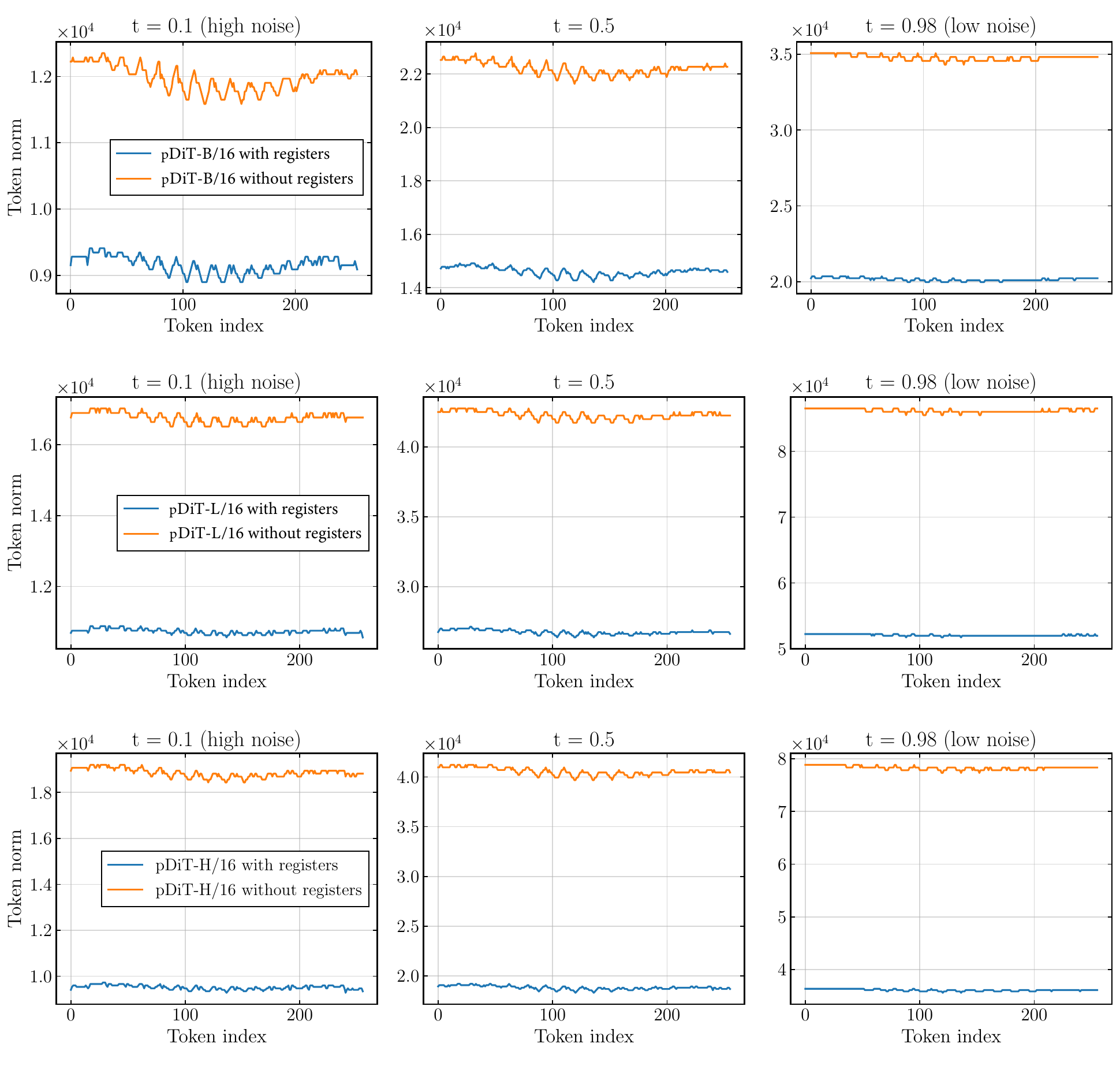}
    \caption{\textbf{Register tokens consistently reduce feature norms across patch tokens.}
We measure feature norms for image tokens only (excluding register tokens)
at three diffusion timesteps for pDiT models of different scales, and
observe a consistent reduction in feature norms across nearly all tokens
when register tokens are used.}
    \label{fig:app_analysis_jit_norms}
\end{figure*}

\begin{figure*}[t!]
    \centering
    \includegraphics[width=1.0\linewidth]{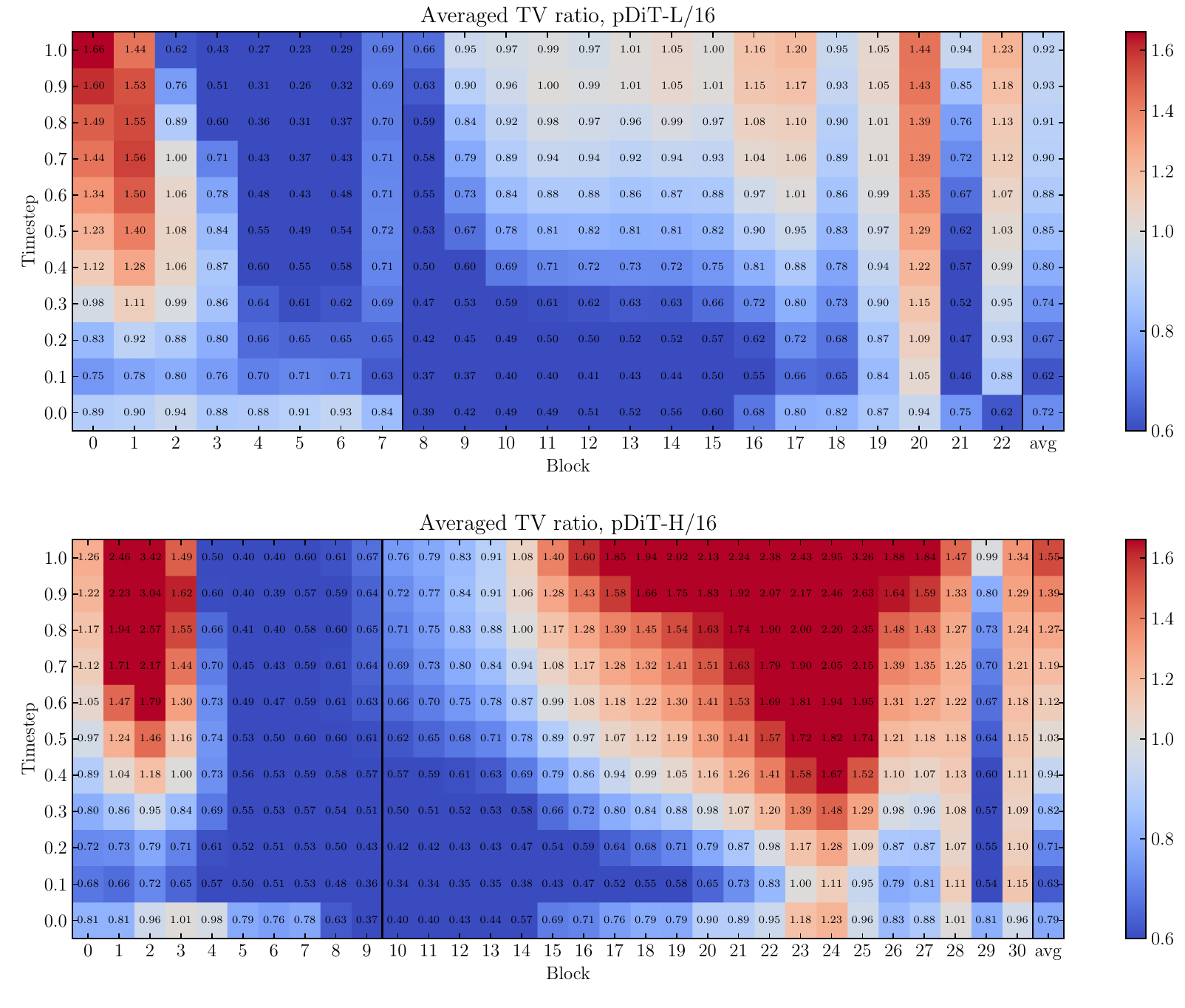}
\caption{\textbf{Register tokens make intermediate representations cleaner by reducing noise.}
We compute the Total Variation of intermediate features for models with
and without register tokens. We report the ratio
(with registers / without registers), where lower values indicate that
models with registers produce smoother feature representations.
We find that register tokens improve feature smoothness at high noise
levels ($t \in [0, 0.2]$) for both pDiT-L/16 (top) and pDiT-H/16 (bottom)
models.}
    \label{fig:app_analysis_jit_tv_ratio}
\end{figure*}

\begin{figure*}[t!]
    \centering
    \includegraphics[width=1.0\linewidth]{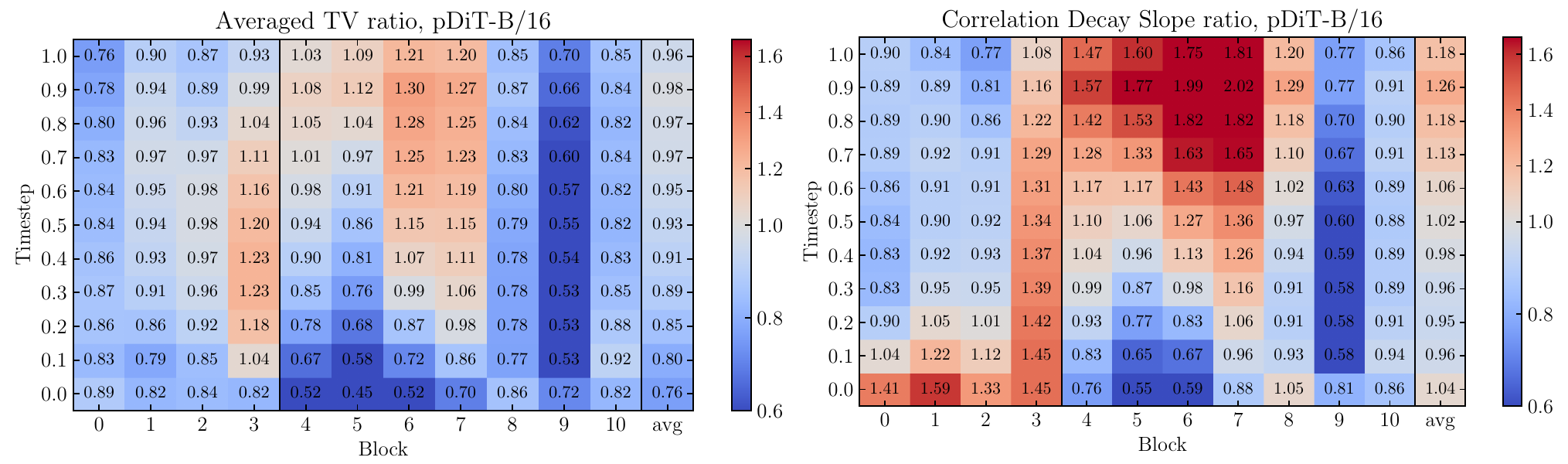}
\caption{\textbf{Registers improve spatial organization at high noise levels.}
In addition to the TV ratio (left), we also analyze the correlation decay slope~\cite{singh2025matters} (right), where lower values indicate stronger spatial organization. 
Both metrics show the same trend: register tokens improve internal representations at high noise levels, starting from block $4$, where the registers are introduced.}
    \label{fig:app_analysis_cka}
\end{figure*}

\begin{figure*}[t!]
    \centering
    \includegraphics[width=1.0\linewidth]{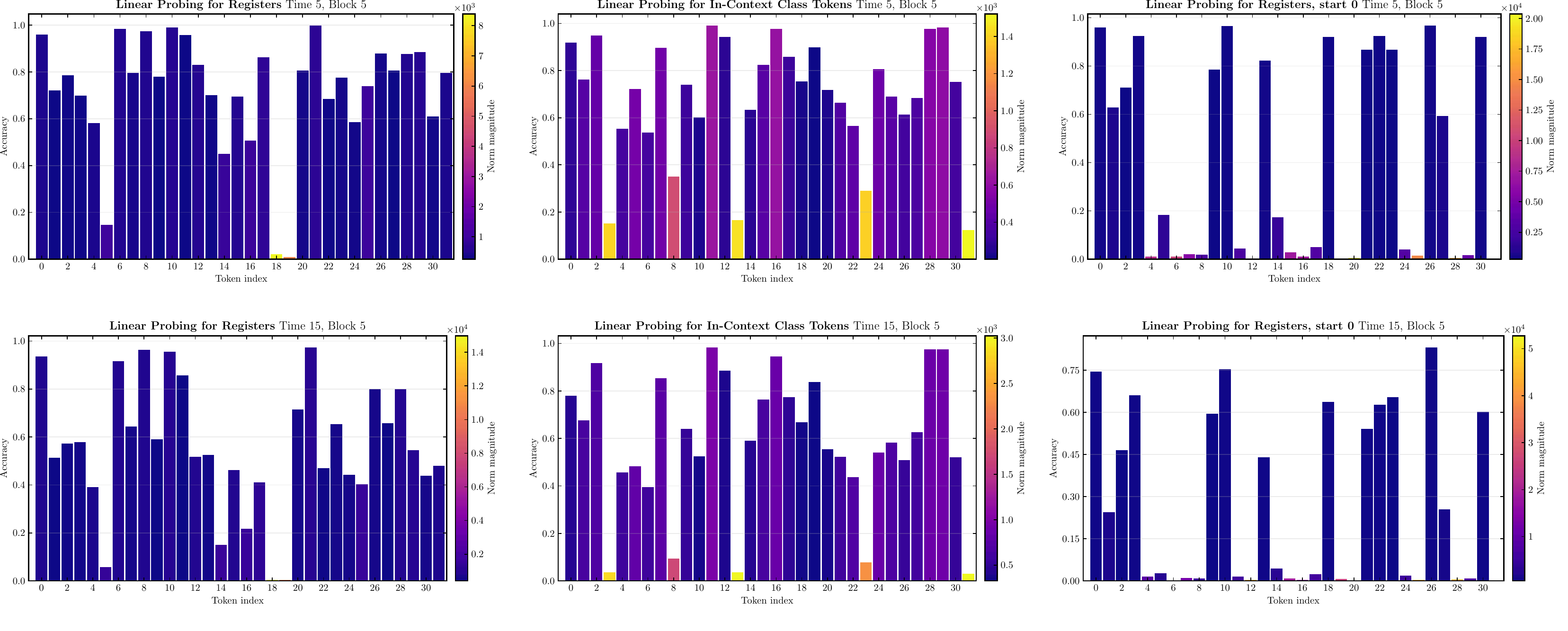}
\caption{\textbf{Linear probing of register tokens under different configurations.} (Left) Standard register tokens introduced from the $4$th layer; (Middle) Register tokens used as in-context class embeddings introduced from the $4$th layer;  (Right) Standard register tokens introduced from the $0$th layer. Across different timesteps, we find that introducing registers from the earliest layers produces substantially less informative register tokens. In particular, we observe more low-norm (non-sink) tokens with poor linear probing accuracy.}
    \label{fig:app_analysis_probing}
\end{figure*}

\begin{figure*}[t!]
    \centering
    \includegraphics[width=1.\linewidth]{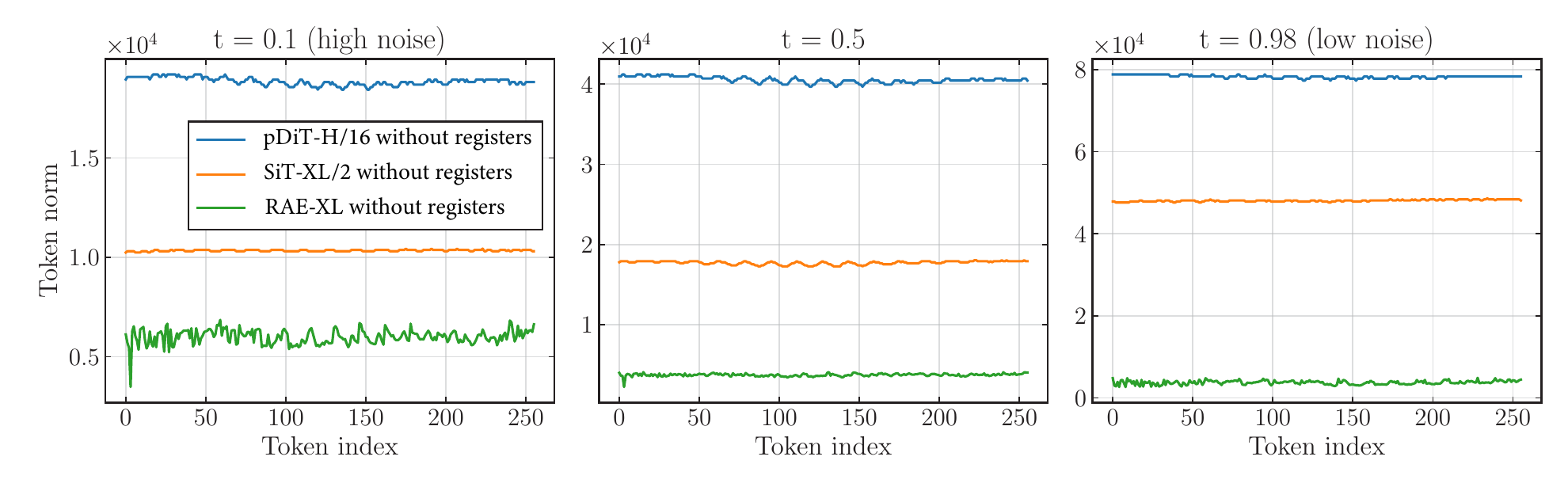}
\caption{
    \textbf{Pixel-space pDiTs have the highest feature norms across all tokens for different timesteps compared to latent-space counterparts.} 
    We compare token-wise feature-map norms for pDiT, SiT, and RAE models, all without register tokens.
}
    \label{fig:app_analysis_norms_models}
\end{figure*}

\begin{figure*}[t!]
    \centering
    \includegraphics[width=1.\linewidth]{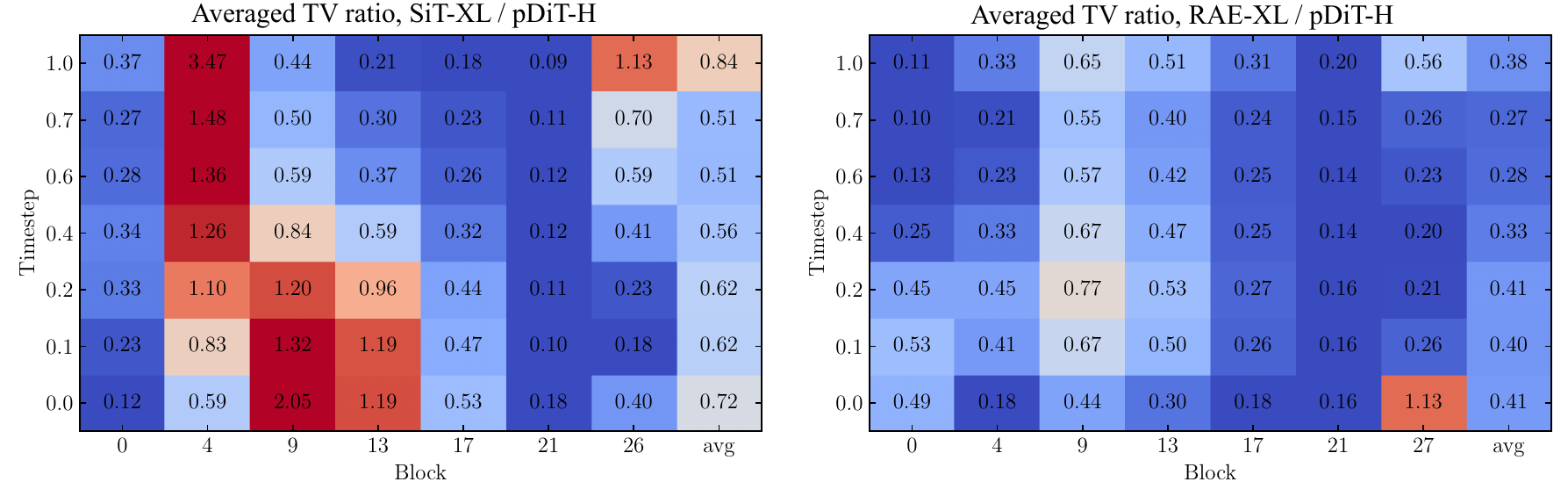}
\caption{\textbf{Pixel-space pDiTs exhibit substantially higher Total Variation (TV) values than latent-space counterparts.} We compare the TV ratio of intermediate feature maps across timesteps and transformer blocks for pixel-space pDiTs (pDiT-H) and latent-space models (SiT-XL and RAE-XL) without registers. Pixel-space pDiTs consistently produce noisier intermediate representations.}
    \label{fig:app_analysis_tvs_models}
\end{figure*}

\begin{figure*}[t!]
    \centering
    \includegraphics[width=1.0\linewidth]{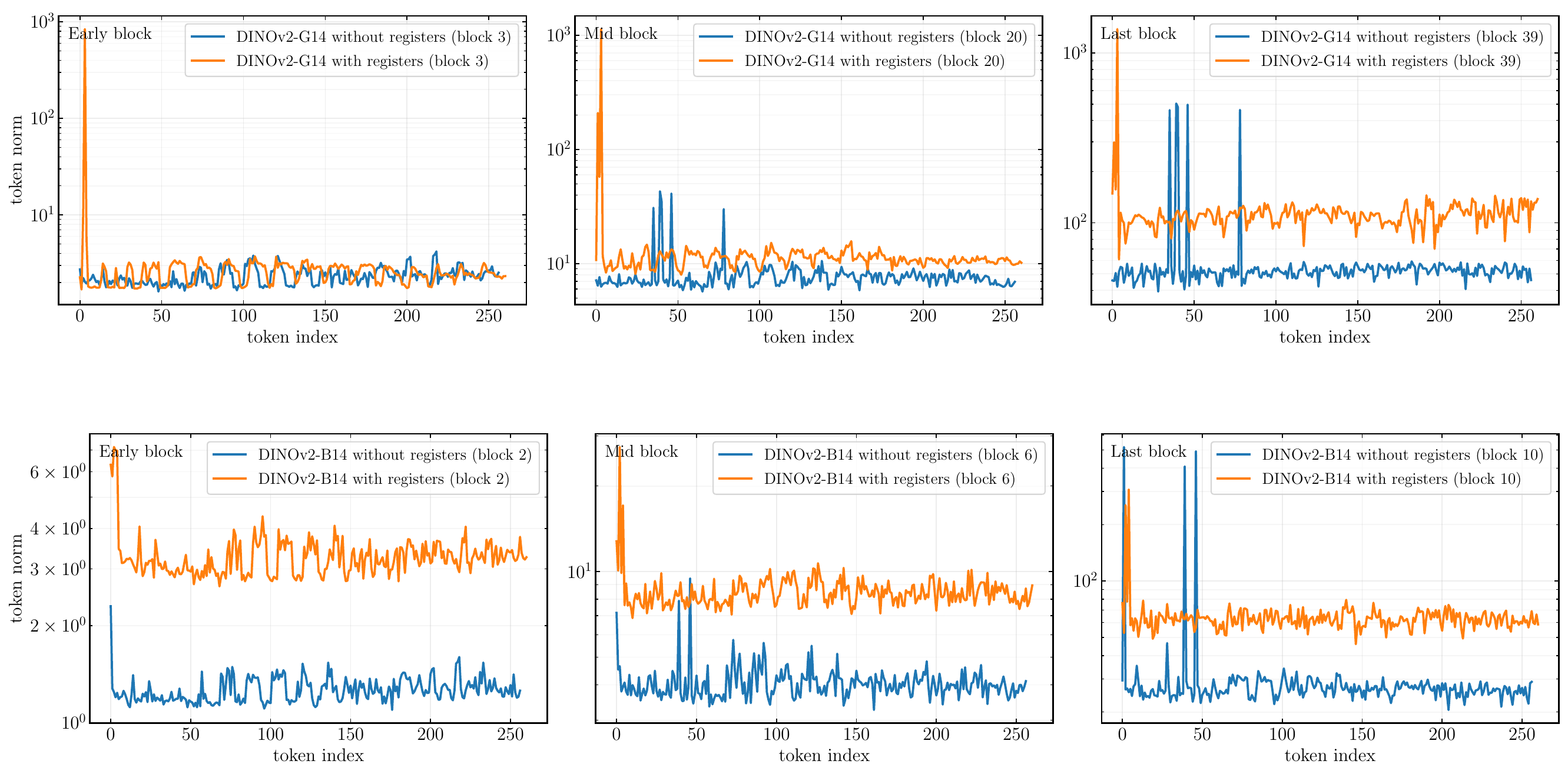}
    \caption{\textbf{For SSL ViTs such as DINOv2, register tokens do not reduce patch-token feature norms, unlike in DiTs.} We measure feature norms across all tokens for different blocks and model sizes of DINOv2, and observe that register tokens do not consistently reduce feature norms of patch tokens.}
    \label{fig:app_analysis_dino_norms}
\end{figure*}

\begin{figure*}[t!]
    \centering
    \includegraphics[width=1.\linewidth]{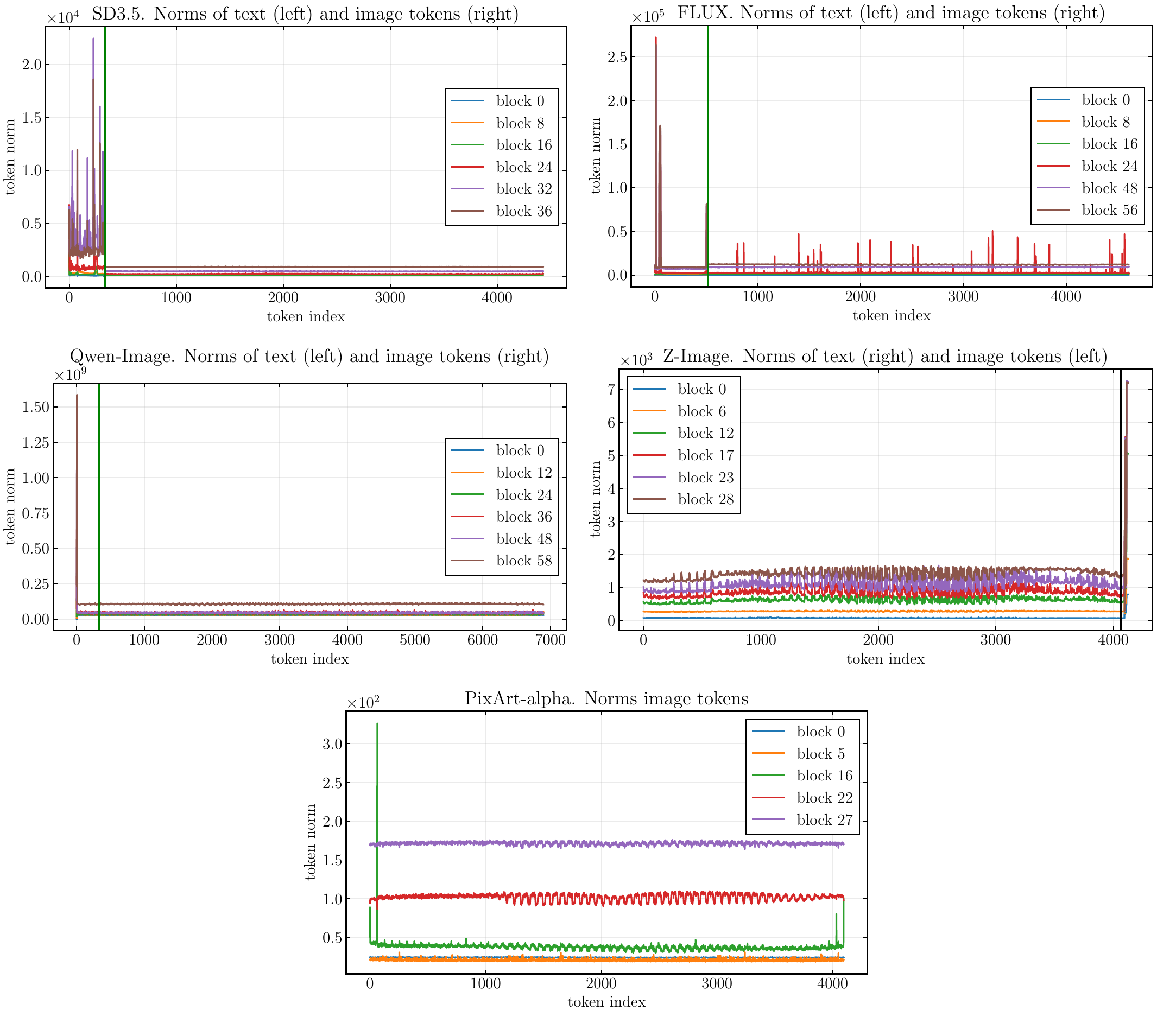}
\caption{\textbf{Text sequences in text-to-image diffusion models exhibit behavior similar to register tokens in ImageNet-based DiTs: some tokens become high-norm outliers and potentially act as registers.} We measure token-wise feature norms in SD3.5 (left) and FLUX (right) for both text and image tokens. We observe that the outliers primarily emerge within the text sequence.}
    \label{fig:app_analysis_t2i}
\end{figure*}

\begin{figure*}
    \centering
    \includegraphics[width=1.0\linewidth]{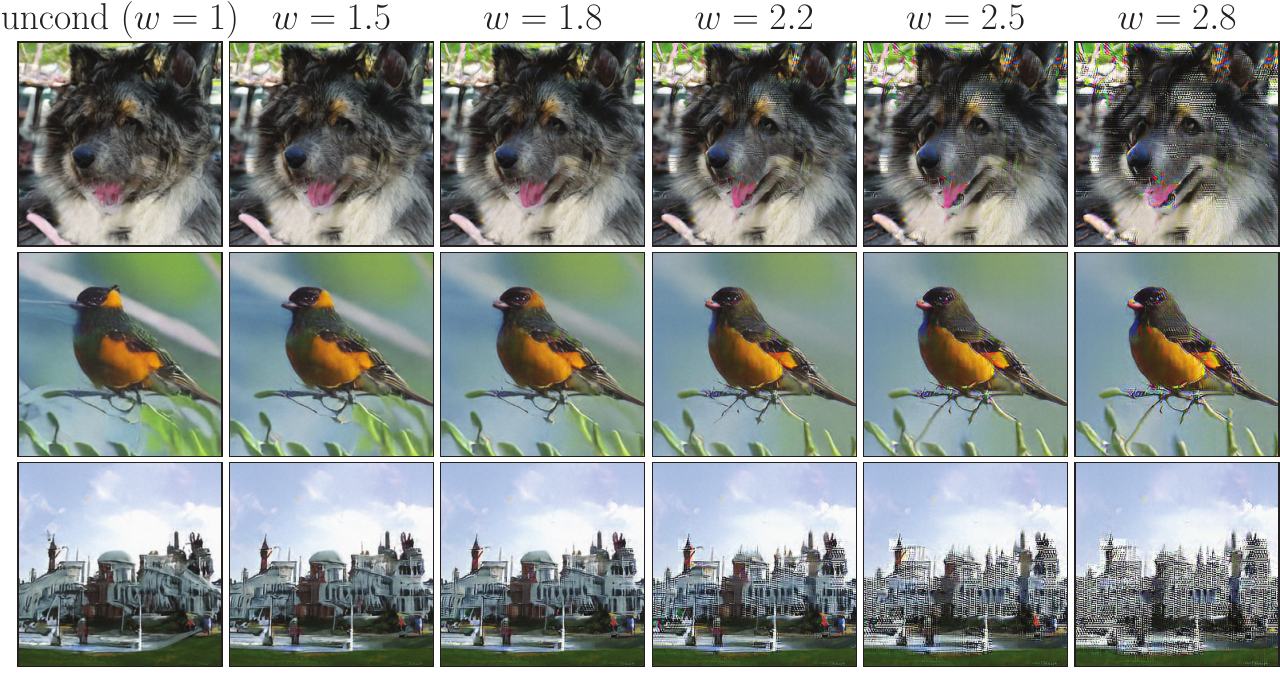}
    \caption{
\textbf{Artifacts from Register Guidance with separate models.}
We apply Register Guidance using predictions from separately trained models with and without registers.
As the guidance scale increases, visible artifacts appear, suggesting imperfect alignment between the two models.
}
    \label{fig:app_uncond_guidance_separate_full}
\end{figure*}
\begin{figure*}
    \centering
    \includegraphics[width=1.0\linewidth]{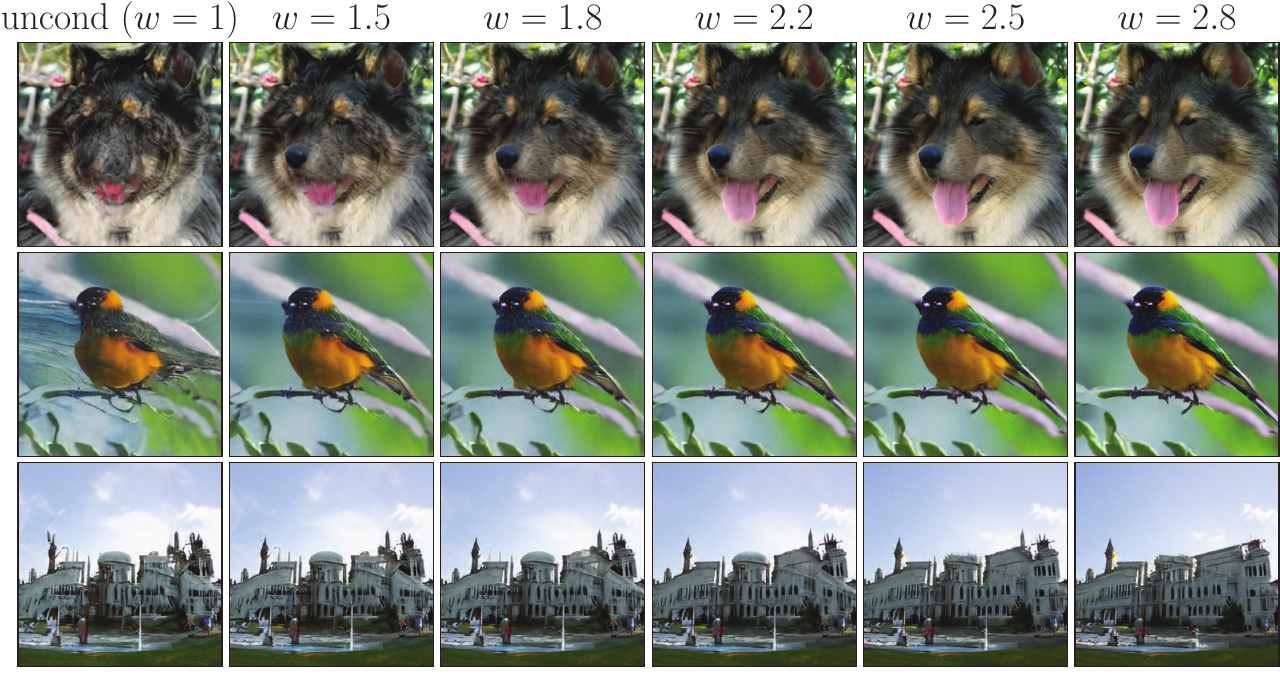}
    \caption{
\textbf{Register Guidance with a single model removes artifacts.}
We apply Register Guidance using a single model trained to operate both with and without registers.
Unlike guidance with separate models, increasing the guidance scale improves structure and details without introducing visible artifacts.
}
    \label{fig:app_uncond_guidance_full}
\end{figure*}

\begin{figure*}[t!]
    \centering
    \includegraphics[width=1.0\linewidth]{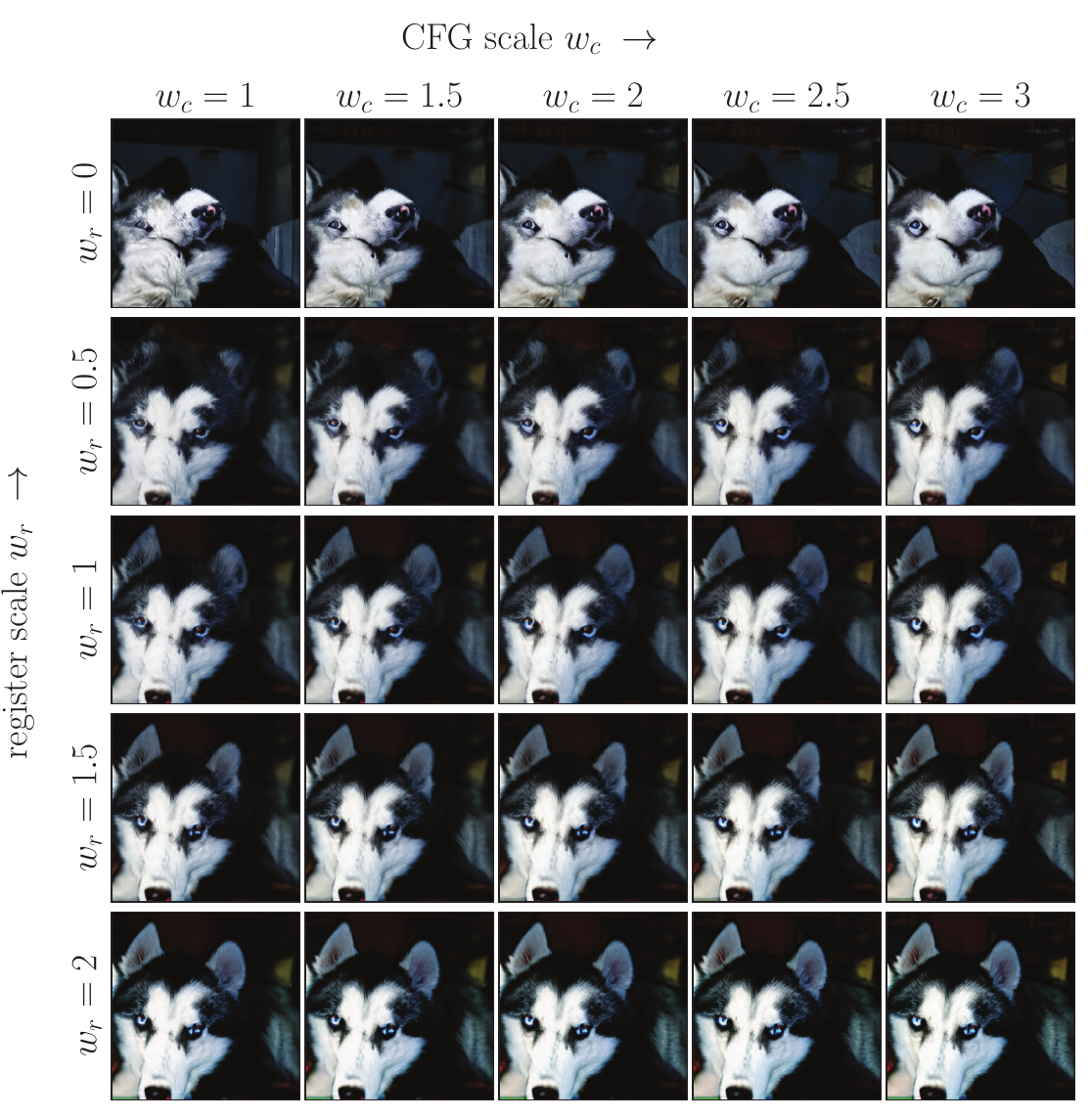}
    \caption{
\textbf{Visual effect of combining CFG and RG.}
We vary the CFG scale $w_c$ horizontally and the RG scale $w_r$ vertically.
CFG strengthens class conditioning, while RG improves structure and visual details, showing that the two guidance signals are complementary.
}
    \label{fig:app_cfg_rg_visual}
\end{figure*}

\begin{figure*}[t!]
    \centering
    \includegraphics[width=1.0\linewidth]{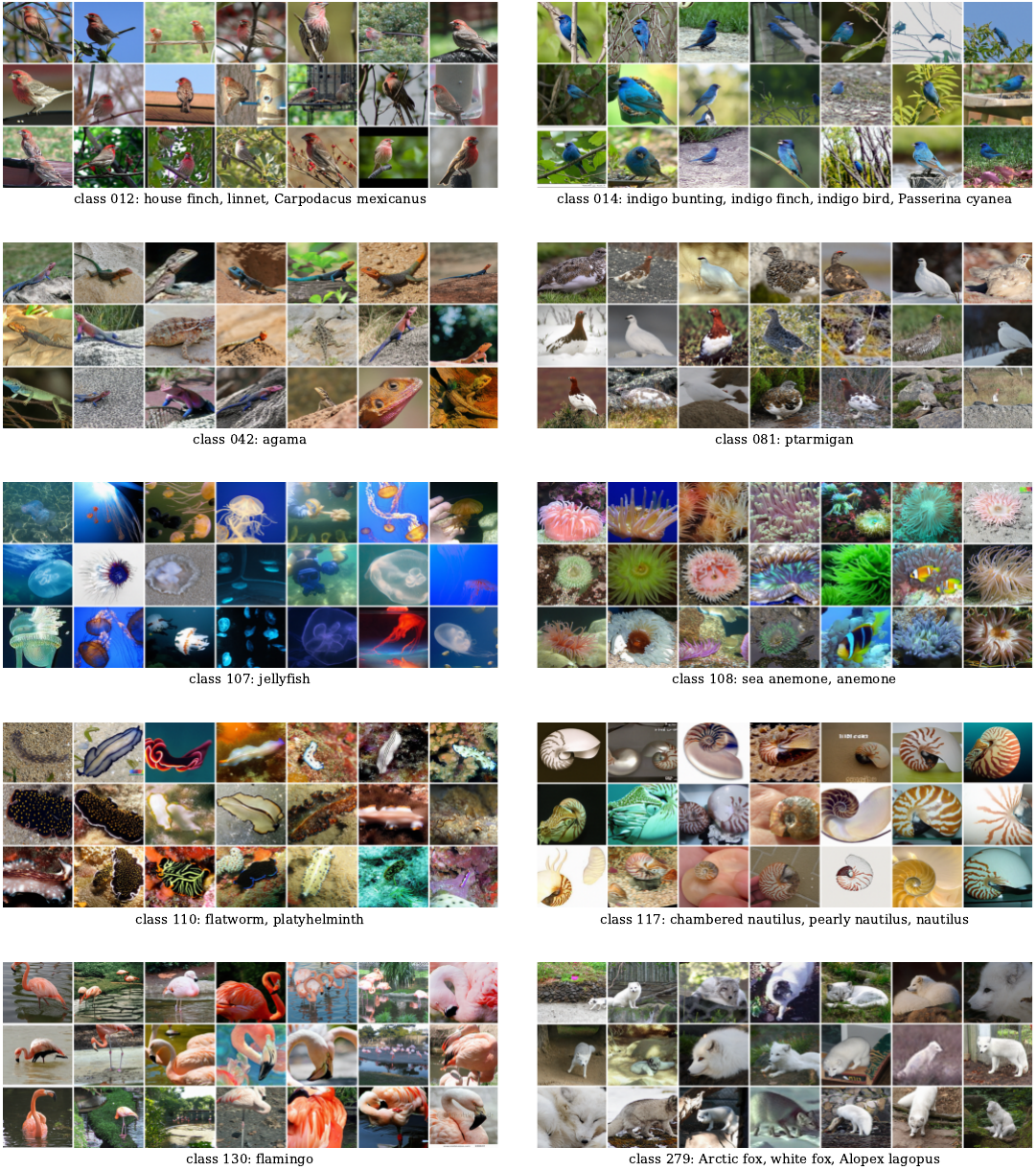}
    \caption{
\textbf{Uncurated samples on ImageNet $256{\times}256$ using JiT-H/16.}
We show images generated with RG+CFG, which achieves the reported FID of $1.80$.
}
    \label{fig:app_rg_cfg_imgs_1}
\end{figure*}

\begin{figure*}[t!]
    \centering
    \includegraphics[width=1.0\linewidth]{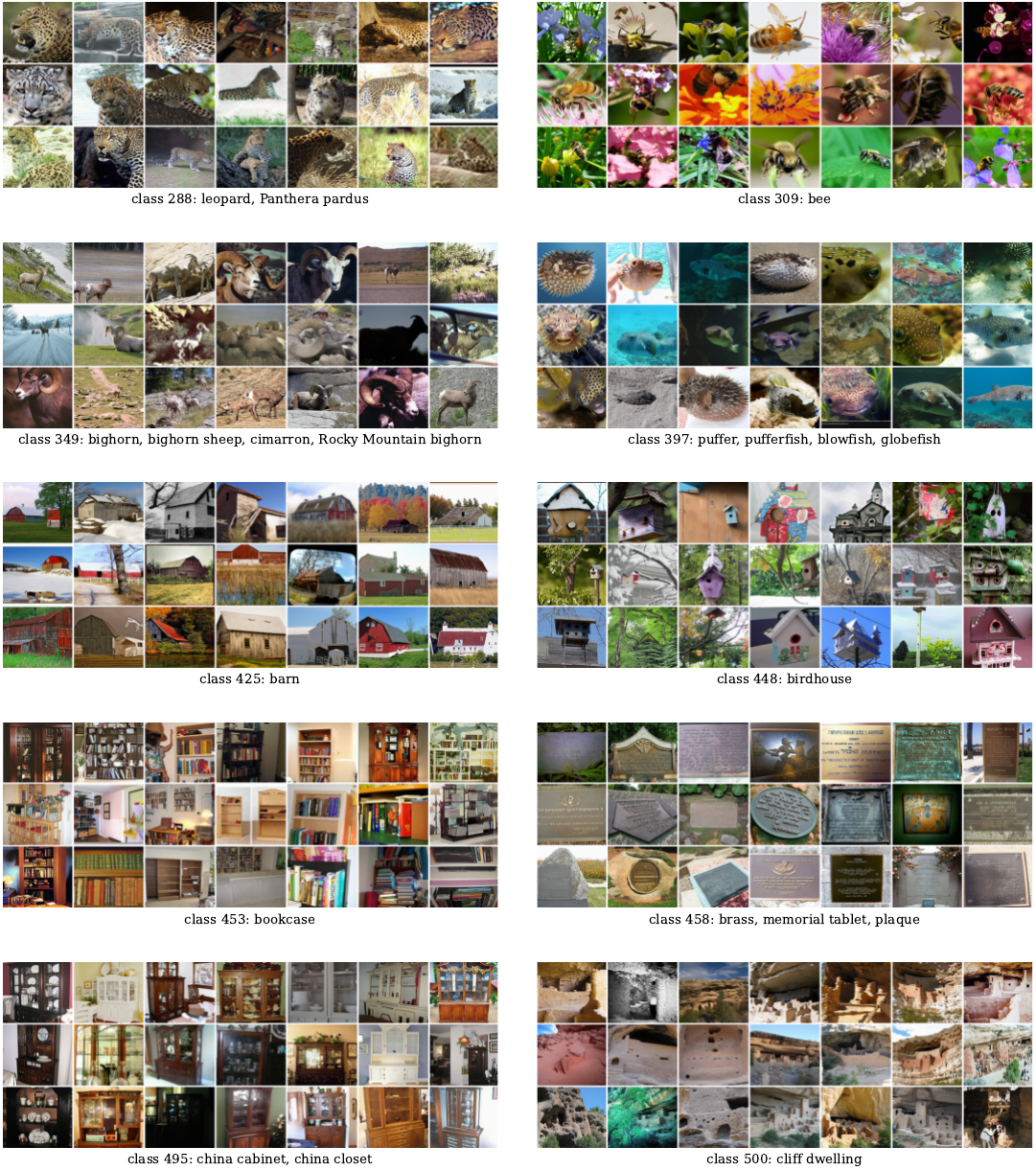}
    \caption{
\textbf{Uncurated samples on ImageNet $256{\times}256$ using JiT-H/16.}
We show images generated with RG+CFG, which achieves the reported FID of $1.80$.
}
    \label{fig:app_rg_cfg_imgs_2}
\end{figure*}


\end{document}